\documentclass[10pt,journal,compsoc]{IEEEtran}

\ifCLASSOPTIONcompsoc
  \usepackage[nocompress]{cite}
\else
  \usepackage{cite}
\fi
\ifCLASSINFOpdf
  \usepackage[pdftex]{graphicx}
\else
  \usepackage[dvips]{graphicx}
\fi
\usepackage[cmex10]{amsmath}
\usepackage{url}

\usepackage{algorithm}
\usepackage{algorithmic}

\usepackage{bm}
\usepackage{subfig}
\usepackage{epsfig}
\usepackage{amssymb}

\newcommand{\argmax}{\operatornamewithlimits{argmax}}
\newcommand{\argmin}{\operatornamewithlimits{argmin}}

\begin{document}

    \title{Image Reconstruction from Bag-of-Visual-Words}
    \author{
        Hiroharu~Kato~and~Tatsuya~Harada
        \IEEEcompsocitemizethanks{
            \IEEEcompsocthanksitem H. Kato and T. Harada are with the Department of Mechano-Informatics, Graduate School of Information Science and Technology, The University of Tokyo.
        }
    }
    \IEEEtitleabstractindextext{
        \begin{abstract}
            The objective of this work is to reconstruct an original image from Bag-of-Visual-Words (BoVW). Image reconstruction from features can be a means of identifying the characteristics of features. Additionally, it enables us to generate novel images via features. Although BoVW is the de facto standard feature for image recognition and retrieval, successful image reconstruction from BoVW has not been reported yet. What complicates this task is that BoVW lacks the spatial information for including visual words. As described in this paper, to estimate an original arrangement, we propose an evaluation function that incorporates the naturalness of local adjacency and the global position, with a method to obtain related parameters using an external image database. To evaluate the performance of our method, we reconstruct images of objects of $101$ kinds. Additionally, we apply our method to analyze object classifiers and to generate novel images via BoVW.
        \end{abstract}

    }

    \maketitle
    \IEEEdisplaynontitleabstractindextext
    \IEEEpeerreviewmaketitle

    \IEEEraisesectionheading{\section{Introduction}}
        \label{chap:introduction}

        \IEEEPARstart{T}{he} most influential factors associated with the performance of an object recognition or retrieval system are image feature characteristics. To develop high-performance systems, we must adopt or develop suitable features for the target task.

        However, few intuitive ways to evaluate the characteristics of image features present themselves. One way is to reconstruct original images from features~\cite{weinzaepfel2011reconstructing,d2012beyond,vondrick2012inverting,zeiler2013visualizing}. By examining reconstructed images, we can elucidate what information the feature captures. This insight is useful for improving the performance of computer vision systems. For example, Vondrick et al.~\cite{vondrick2012inverting} analyzed classification failures of HOG features through image reconstruction and found weaknesses of HOG. The winners of ILSVRC $2013$ object recognition competition~\cite{ILSVRCarxiv14} carefully tuned the hyperparameters of their system by visualizing learned features~\cite{zeiler2013visualizing}.

        Moreover, image reconstruction has another application: image generation via features. Image feature represents the semantics of images. Therefore feature-based image generation can be a mode of semantic image modification or cross-modal translation. For example, Chen et al.~\cite{chen2014inferring} inferred an unseen view of people and generated images via the HOG feature. Although content generation via features might sound strange in computer vision, it is a commonly used approach for audio signal processing and other activities. The most popular audio feature, MFCC, is widely used as intermediate representation for speech synthesis~\cite{zen2009statistical} and voice conversion~\cite{stylianou1998continuous}.

        \begin{figure}[t]
            \begin{center}
                \includegraphics[bb=0 0 609 601,width=1.0\linewidth]{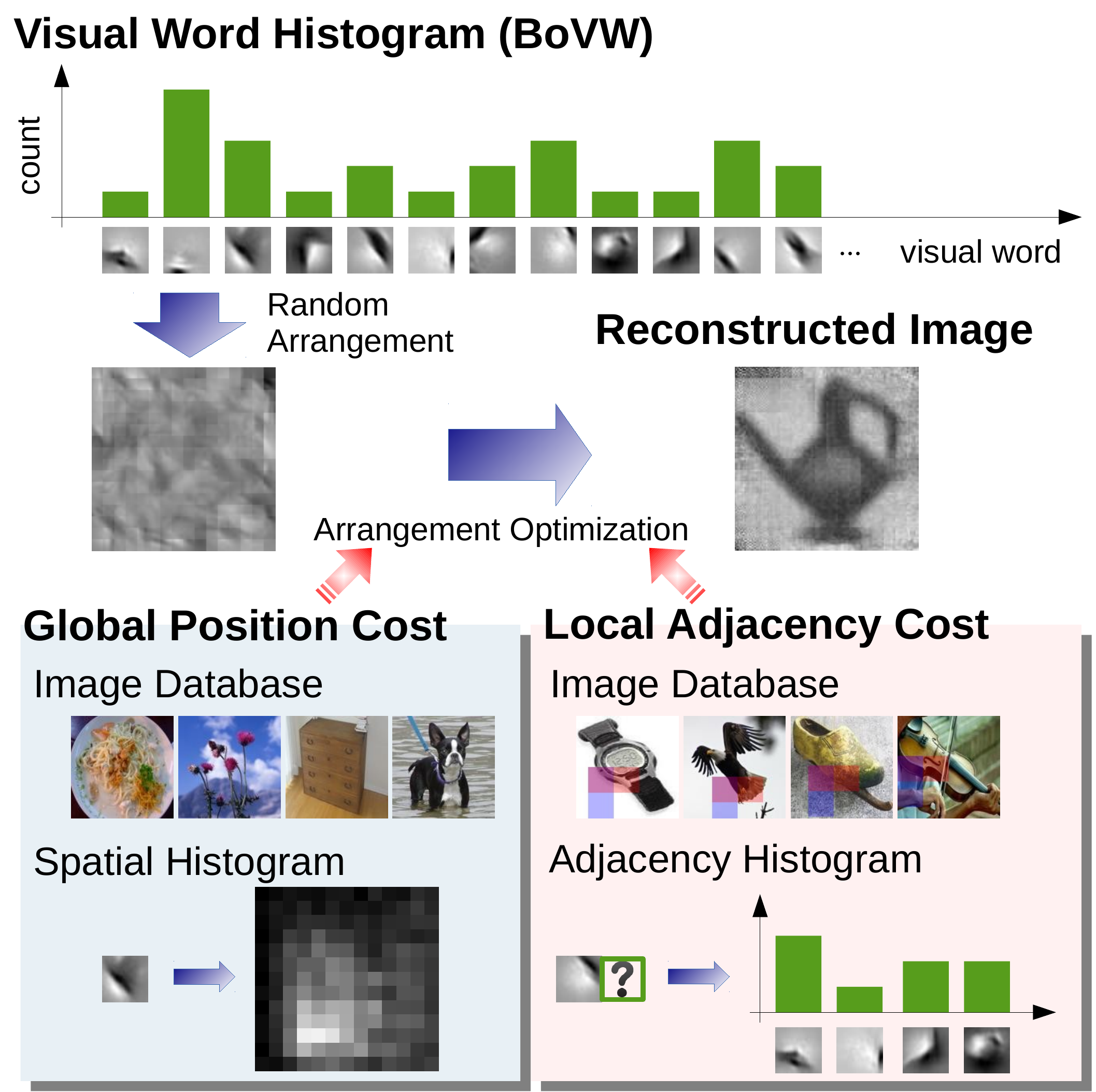}
            \end{center}
            \caption{Illustration of our reconstruction method. First, the spatial arrangement of visual words is optimized by maximizing the naturalness of local adjacency and the global position. Then each visual word is converted and blended into a single image.}
            \label{fig:system}
        \end{figure}

        Bag-of-Visual-Words (BoVW), the de facto standard feature for image recognition and retrieval, is the basis of many modern features \cite{sanchez2013image, zhou2010image}. However, no method of reconstructing images from BoVW has been reported in the literature. Therefore, we developed such a method in this work.

        BoVW is a set of local descriptors extracted from the local regions of an image. Existing works~\cite{weinzaepfel2011reconstructing,d2012beyond,vondrick2012inverting} have demonstrated the capability of to converting local descriptors into original image patches. However, arranging them is difficult because BoVW lacks their related spatial information. Therefore, we must recover the original layout, which is the main point of this work. Fig.~\ref{fig:system} presents our method. We re-arrange local descriptors by maximizing the naturalness of local adjacency and the global position like solving jigsaw puzzles. This produces an optimization problem called the Quadratic Assignment Problem. We solve it using a generic solver based on meta-heuristics.

        The contributions of this paper are summarized as follows. 1) We tackled a novel problem of reconstructing images from BoVW. 2) We decomposed this problem into estimation of the optimal layout of local descriptors and proposed an evaluation function that measures the naturalness of layouts using an external image database. 3) We described the relation among our evaluation function, jigsaw puzzle solvers, and Quadratic Assignment Problem. 4) We applied our method to analyze object classifiers and to generate novel images from natural sentences.

        A preliminary version of this work has been presented elsewhere~\cite{Kato_CVPR_2014}. This presentation includes the addition of three image-generation applications as described in Section \ref{chap:application}, along with evaluation by humans, and more detailed discussion. Our proposed method includes slight modifications.

    \section{Related Work}
        \label{chap:related_work}

        In this section, we summarize existing image reconstruction methods and explain the relations between ours and the others. Additionally, we describe existing approaches of image generation.

        \subsection{Image Reconstruction}
            \label{sec:related_work_reconstruction}
            
            The work by Weinzaepfel et al.~\cite{weinzaepfel2011reconstructing} is the first to reconstruct images from features. They used SIFT descriptors~\cite{lowe2004distinctive} with geometric information for reconstruction. Because they sampled descriptors at keypoints, the geometric information includes the size, orientation, and position of the image patch from which the descriptor is extracted. They construct a database of local descriptors. Their corresponding image patches are taken from an external image database in advance. To reconstruct an image, they retrieve the most resemblant image descriptor in the database for each local descriptor, transform their corresponding image patches according to their geometric information, and blend them by Poisson Blending~\cite{perez2003poisson}. They assumed that geometric information of local descriptors is available, which is rarely true in realistic systems.

            d'Angelo et al.~\cite{d2012beyond} proposed a method to convert BREAF descriptors~\cite{calonder2010brief} and FREAK descriptors~\cite{alahi2012freak} into image patches. Their method generates an optimal image patch for the input descriptor analytically. Although the generated images are clear, their method is applicable only to  descriptors of certain kinds.

            Vondrick et al.~\cite{vondrick2012inverting} reconstructed images from HOG features~\cite{dalal2005histograms}. They proposed four methods and concluded that an approach that learns pairs of an image features and the corresponding image patch by Sparse Coding is effective. Their method is applicable to any feature, in principle.

            Visualization of features is also important for representation learning. For widely used Deep Convolutional Neural Networks (CNN) \cite{lecun1998gradient}, loss of spatial information at the pooling step is problematic, as it is with BoVW. Zeiler et al.~\cite{zeiler2013visualizing, zeiler2010deconvolutional} addressed this problem by saving spatial information at the pooling step. Mahendra et al.~\cite{mahendran2014understanding} inverted CNNs by generating an optimal image for the feature directly using Stochastic Gradient Descent.

            Except for representation learning, all methods described above are useful for converting local descriptors into image patches. However, BoVW includes many quantized local descriptors without their geometric information. Our work is the first to address such a feature.
            
        \subsection{Image Generation}
            \label{sec:related_work_generation}
           
            A conventional approach for image generation is to synthesize multiple images~\cite{johnson2006semantic, chen2009sketch2photo} because direct generation of images is rather difficult. They can produce clear and aesthetic images. However, they need an extremely large labeled and segmented image dataset, which entails high costs. Moreover, they can only produce a patchwork of existing images; they cannot produce truly novel images.
            
            With recent rapid progress in Neural Networks, they have started to be used for image generation. Generative models such as Deep Belief Networks and Variational Auto-Encoders can generate images directly from features~\cite{erhan2009visualizing, bengio2012better, kingma2013auto}. Generative properties of Recurrent Neural Networks are also used for generation~\cite{graves2013generating, gregor2015draw}. Although their potential versatile capacities are attractive for image generation, for now they are limited to certain domains such as handwriting digits or characters, faces, and the CIFAR-10 dataset~\cite{krizhevsky2009learning}.

    \section{Reconstruction Method}
        \label{chap:method}

        In this section, we propose a method to reconstruct images from BoVW. First, we provide a brief introduction to BoVW and specify the definitions of BoVW in this paper. Then we define our problem setting and decompose our problem into estimation of the spatial layout of visual words and generation of images from them. For the former, we show similarity to jigsaw puzzle solvers, propose an evaluation function, and describe the optimization method. We also describe a means to evaluate the reconstruction performance.

        \subsection{Bag-of-Visual-Words}
            \label{sec:related_work_bovw}

            BoVW is the de-facto standard feature for image retrieval~\cite{sivic2003video} and recognition~\cite{csurka2004visual}. It was inspired by Bag-of-Words (BoW) proposed in the natural language processing community. BoW is defined as a histogram of words included in a document. Similarly, BoVW is defined as a histogram of visual words included in an image.

            The basic procedure to extract BoVW from an image is the following. 1) The first step is called {\it sampling}. In this step, descriptors such as SIFT \cite{lowe2004distinctive} are extracted from small local regions of the image. 2) The next step is called {\it coding}. In this step, each local descriptor is assigned to the nearest visual word. Visual words are obtained as centroids of clusters of local descriptors extracted from an image dataset. 3) The last step is called {\it pooling}. In this step, a histogram of visual words is computed by counting how many descriptors are assigned to each visual word.

            Two major strategies exist for the sampling step. One is to extract descriptors from local image patches detected by keypoint detectors~\cite{harris1988combined, lowe2004distinctive}; the other is from fixed grid points. The size and orientation of local image patches are variable at the former one and fixed at the latter one. The latter, designated as {\it dense sampling}, is better for image recognition and retrieval~\cite{nowak2006sampling, gordoa2012leveraging}.

            Several options exist for the coding step. Instead of assigning each local descriptor to one visual word ({\it hard assignment}), some method~\cite{winn2005object, perronnin2006adapted, van2008kernel, yang2009linear, wang2010locality, boureau2010learning} represents each descriptor by a vector, the $i$-th dimension of which is the weight for the $i$-th visual word.

            The pooling method described above is designated as {\it sum pooling} because it sums up assigned descriptors of each visual word. Some works~\cite{yang2009linear, boureau2010theoretical} use {\it max pooling}, which takes the max instead of the sum.

            Many studies have been undertaken to embed spatial information of local descriptors into features. The most famous of them is the Spatial Pyramid~\cite{lazebnik2006beyond}.

            Actually, BoVW has been extended to state-of-the-art hand-engineered features, such as VLAD~\cite{jegou2010aggregating}, Super Vector~\cite{zhou2010image}, and Fisher Vector~\cite{sanchez2013image}. Although their recognition performance is superior to that of BoVW, the differences are not great~\cite{chatfield2011devil}; BoVW is still widely used.

            In this work, we assume BoVW as the simplest style. We use dense sampling for the sampling step, hard assignment for the coding step, and sum pooling for the pooling step. We do not use any method to embed the spatial information of local descriptors.
            
        \subsection{Problem Settings}
            \label{sec:method_settings}

            Three approaches are possible for image reconstruction. The first is to compute optimal images analytically as in the approach described by d'Angelo et al.~\cite{d2012beyond}, which is difficult for BoVW because BoVW is extracted through highly complicated transformations. The second is to learn the correspondence between features and images directly from an external image database. We evaluate this approach by application of a method by Vondrick et al.~\cite{vondrick2012inverting} in the experiment section of this report. The third is to trace back the pipeline of extraction of BoVW. We particularly study this approach in this paper.

            For this work, we assume dense sampling, hard assignment, and sum pooling for BoVW, as described above. Additionally, we assume the availability of the dictionary of visual words, grid size of dense sampling, and the original image size.

            In this setting, we can ascertain the number and the kind of visual words included in an image from BoVW. Missing geometric information of visual words is only the positions in the image because the size and orientation of image patches are fixed at dense sampling. To trace back the pipeline of BoVW, it is necessary to estimate the original arrangement of visual words (for pooling) and to generate image patches from them (for sampling). We assume that the quantization error at the coding step is negligible.

        \subsection{Analogy to Jigsaw Puzzles}
            \label{sec:method_analogy_with_jigsaw}

            Because a set of visual words and a grid structure are given, recovery of their spatial arrangement  is to ascertain the optimal assignments of $N$ visual words to $N$ grid points. This problem is similar to solving a jigsaw puzzle of $N$ square pieces without the original image, which has been studied recently~\cite{cho2010probabilistic, pomeranz2011fully, gallagher2012jigsaw, sholomon2013genetic}.

            Strategies of two kinds are possible to solve a jigsaw puzzle. One is to consider the naturalness of the local adjacency of pieces. Images of adjacent pieces must be smooth and continuous. The other is to consider the naturalness of the global structure. For example, pieces of sky tend to be situated at the top of the image. These naturalnesses are evaluated according to our past experiences of seeing images.

            From these insights, to rearrange visual words, we make their local adjacency and global position natural.

        \subsection{Evaluation Function}
            \label{sec:method_cost_function}

            An assignment of $N$ elements to $N$ places can be represented by a permutation matrix $x$. $x_{ik}$ is $1$ only if the $i$-th element is assigned to $k$-th place; otherwise $0$. Actually, $x$ has the following constraints so that multiple elements are not assigned to one place.
            \begin{align}
                \sum_{i=1}^{N} x_{ik} = 1 \;\;\;\; & \left( 1 \leq k \leq N \right). \label{eq:st1}\\
                \sum_{k=1}^{N} x_{ik} = 1 \;\;\;\; & \left( 1 \leq i \leq N \right). \label{eq:st2}\\
                x_{ik} \in \{ 0,1 \} \;\;\;\; & \left( 1 \leq i,k \leq N  \right). \label{eq:st3}
            \end{align}

            Let $C_{ijkl}^{a}$ represent the unnaturalness of an assignment by which $i$-th and $j$-th elements are placed respectively at $k$-th and $l$-th places. $C_{ijkl}^{a}$ is not $0$ only if $k$-th and $l$-th places are adjacent. We designate $C^{a}$ as {\it Local Adjacency Cost}. Then the sum of the Local Adjacency Cost of all elements $E^{a}(x)$ is the following.
            \begin{align}
                E^{a}(x) = \sum_{i,j,k,l=1}^{N} {C^{a}_{ijkl} x_{ik} x_{jl}} \
            \end{align}

            Let $C_{ik}^{p}$ represent the unnaturalness of an assignment by which $i$-th element is assigned to the $k$-th place. We designate $C^{p}$ as the {\it Global Position Cost}. Then the sum of the Global Position Cost of all elements $E^{p}(x)$ is the following.
            \begin{align}
                E^{p}(x) = \sum_{i,j=1}^{N} C_{ik}^{p} x_{ik} \
            \end{align}

            Our evaluation function $E(x)$ is defined as the weighted sum of both unnaturalnesses. Let $\lambda \; (0 \leq \lambda \leq 1)$ as weighting parameter. $E(x)$ is defined as explained below.
            \begin{align}
                E(x) = (1-\lambda) E^{a} (x) + \lambda E^{p} (x). \label{eq:cost}
            \end{align}
            We obtain optimal arrangement $x^*$ by solving the following optimization problem.
            \begin{align}
                x^* =& \argmin E(x) \; . \nonumber \\
                \mbox{s.t.} \;\;\ & \mbox{Equation} \;~\ref{eq:st1}, \;~\ref{eq:st2}, \;~\ref{eq:st3}. \label{eq:problem}
            \end{align}

        \subsection{Local Adjacency Cost}
            \label{sec:method_adjacency_cost}

            Local Adjacency Cost $C^a$ gives reconstructed images consistent edges and shapes. Measuring the naturalness of adjacent visual words is rather more difficult than measuring adjacent jigsaw pieces because visual words have no apparent edge or shape. Additionally, the way to decide $C^a$ is preferably independent of the kind of descriptors. Therefore, we construct parameters of $C^a$ using statistics from an external image database.

            We obtain $C^a$ as co-occurrence statistics of visual words in local regions of images. First, we extract descriptors densely from an image database and quantize them to visual words, saving their geometric information. Second, for all possible neighboring patterns and all possible kinds of visual word pairs, we count the pairs exist in the database. We only consider the relative position of two visual words and neglect their absolute position. We also dismiss pairs for which the distance is more than $m$-neighbor. We set $m=48$, which means we consider $7 \times 7$ elements centered on a certain element. Finally, we add $1$ to all counts for smoothing, normalize them, and take their negative logarithm. Algorithm~\ref{alg:adjacency_cost} shows our method to obtain $C^a$. This procedure is similar to learning a uni-gram language model from a text corpus. The bottom-right part of Fig.~\ref{fig:system} is an illustration of this algorithm.

            \begin{algorithm}[t]
                \caption{Determination of Local Adjacency Cost.}
                \begin{algorithmic}
                    \REQUIRE{image database $I$, distance parameter $m$ }
                    \ENSURE{Local Adjacency Cost $C^{a}$}
                    \STATE{initialize $C^{a'}$ with zeros}
                    \STATE{$N \leftarrow $number of positions in an image}
                    \FOR{\textbf{each} image in $I$}
                        \FOR{\textbf{each} position $k$, $l$ in the image}
                            \IF{$k$ is within $m$-neighbor to $l$}
                                \STATE{$w_k \leftarrow $ visual word extracted at $k$}
                                \STATE{$i \leftarrow $ visual word number of $w_k$}
                                \STATE{$w_l \leftarrow $ visual word extracted at $l$}
                                \STATE{$j \leftarrow $ visual word number of $w_l$}
                                \STATE{$d \leftarrow $ relative position of $k$ and $l$}
                                \STATE{$C^{a'}_{ijd} \leftarrow C^{a'}_{ijd}+1 $}
                            \ENDIF
                        \ENDFOR
                    \ENDFOR
                    \STATE{$C^{a'} \leftarrow C^{a'} + 1$}
                    \FOR{\textbf{each} relative position $d$}
                        \FORALL{$i$ such that $1 \leq i \leq N$}
                            \STATE{normalize $C^{a'}_{ij d}$ such that $\sum_{j} C^{a'}_{ij d} = 1$}
                        \ENDFOR
                    \ENDFOR

                    \FORALL{$i,j,k,l$ such that $1 \leq i,j,k,l \leq N$}
                        \STATE{$d \leftarrow $ relative position of $k$ and $l$}
                        \STATE{$C^{a}_{ijkl} \leftarrow - \log \left( C^{a'}_{ijd} \right) $}
                    \ENDFOR
                    \RETURN{$C^{a}$}
                \end{algorithmic}
                \label{alg:adjacency_cost}
            \end{algorithm}

        \subsection{Global Position Cost}
            \label{sec:method_position cost}

            \begin{algorithm}[t]
                \caption{Determination of Global Position Cost.}
                \begin{algorithmic}
                    \REQUIRE{image database $I$}
                    \ENSURE{Global Position Cost $C^{p}$}
                    \STATE{initialize $C^{p}$ with zeros}
                    \FOR{\textbf{each} image in $I$}
                        \FOR{\textbf{each} position $k$ in the image}
                            \STATE{$w \leftarrow $ visual word extracted at $k$}
                            \STATE{$i \leftarrow $ visual word number of $w$}
                            \STATE{$C^{p}_{ik} \leftarrow C^{p}_{ik}+1 $}
                        \ENDFOR
                    \ENDFOR
                    \STATE{$C^{p} \leftarrow C^p + 1$}
                    \FOR{\textbf{each} visual word number $i$ in the dictionary}
                        \STATE{normalize $C^{p}_{ik}$ such that $\sum_{k} C^{p}_{ik} = 1$}
                    \ENDFOR
                    \STATE{$C^{p} \leftarrow - \log \left( C^{p} \right) $}
                    \RETURN{$C^{p}$}
                \end{algorithmic}
                \label{alg:position_cost}
            \end{algorithm}

            The Global Position Cost $C^p$ makes reconstructed images globally feasible. Each visual word is presumed to have a preference of the absolute position to be placed at. This preference cannot be obtained from the visual word itself. Therefore we use an external image dataset.

            We obtain $C^p$ as occurrence statistics of a certain visual word at a certain place. First, we extract descriptors densely from an image database and quantize them to visual words. Second, for all possible places and all possible kinds of visual word, we count up the visual words existing in the database. Finally, we add $1$ to all counts for smoothing, normalize them and take their negative logarithm. Algorithm~\ref{alg:position_cost} presents our method to obtain $C^a$. The bottom-left part of Fig.~\ref{fig:system} is an illustration of this algorithm.

        \subsection{Optimization}
            \label{sec:method_optimization}

            Solving Equation~\ref{eq:problem} is not straightforward. The number of possible solutions is $N!$. Therefore, it is unrealistic to solve it using brute-force algorithms. In this section, we show that this problem can result in optimization of Markov Random Field and Quadratic Assignment Problem, and propose an optimization method based on meta-heuristics.

            \subsubsection{Relation to Markov Random Fields}
                The energy function of Markov Random Fields is given as follows.
                \begin{align}
                    E(l) = \sum_{k \in \cal{P}} D_k (l_k) + \sum_{k,l \in \cal{N}} V_{k,l} (l_k, l_l) \; . \label{eq:mrf}
                \end{align}
                Therein, $l_k$ is a label assigned to $k$-th node. $\cal{P}$ is a set of all nodes and $\cal{N}$ is a set of pairs of nodes which affect each other. $D_k (l_k)$ is the preference of each node of each label. $V_{k,l} (l_k,l_l)$ represents effects of the connected nodes. For example, in noise reduction of images, $D_k (l_k)$ stands for the cost to change the pixel value from the original value; $V_{k,l} (l_k,l_l)$ denotes the cost for smoothness of the neighboring pixels.

                By defining $D_k (l_k)$ and $V_{k,l} (l_k,l_l)$ as follows, our evaluation function Equation~\ref{eq:cost} results in the energy function Equation~\ref{eq:mrf}.
                \begin{align}
                    & D_k (l_k) = (1-\lambda) \sum_{i=1}^N C_{ik}^p x_{ik}  \; . \\
                    & V_{k,l} (l_k, l_l) = \lambda \sum_{i,j=1}^N C_{ijkl}^a x_{ik} x_{jl}  \; .
                \end{align}

                Optimization of Markov Random Fields is known as NP-hard. However, leaving restrictions Equation~\ref{eq:st1},~\ref{eq:st2},~\ref{eq:st3} out of consideration, it can be solved approximately by belief propagation~\cite{pearl1988probabilistic}, $\alpha$-expansion~\cite{boykov2001fast}, or sequential tree-reweighted message passing~\cite{kolmogorov2006convergent}, which are not applicable under the restrictions Equation~\ref{eq:st1},~\ref{eq:st2},~\ref{eq:st3}. Cho et al.~\cite{cho2008patch} solved the same problem as ours by introducing an approximation to belief propagation. We tried their method in our setting. However, we obtained an invalid solution because their approximation is too coarse for our problem.

            \subsubsection{Relation to Quadratic Assignment Problem}
                Within problems of assigning $N$ elements to $N$ places, ones objective function of that are quadratic functions is called Quadratic Assignment Problem (QAP)~\cite{koopmans1957assignment}. Among the several definitions of the objective function of QAP, a formulation by Lawler~\cite{lawler1963quadratic}, defined as follows, is the most general form.
                \begin{align}
                    \min           & \sum_{i,j,k,l=1}^{N} c_{ijkl} x_{ik} x_{jl} \label{eq:qap} \; . \\
                    \mbox{s.t.} \; & \mbox{Equation} \;~\ref{eq:st1}, \;~\ref{eq:st2}, \;~\ref{eq:st3} \; .
                \end{align}
                Our objective function Equation~\ref{eq:cost} can be transformed as follows.
                \begin{align}
                    \sum_{i,j,k,l=1}^{N} \left( (1-\lambda) C^{a}_{ijkl} + \frac{\lambda}{N^2} C^{p}_{ik} \right) x_{ik} x_{jl} \; .
                \end{align}
                Therefore, we can solve our optimization problem as QAP by defining $c$ as follows.
                \begin{align}
                    c_{ijkl} = \left( (1-\lambda) C^{a}_{ijkl} + \frac{\lambda}{N^2} C^{p}_{ik} \right) \; .
                \end{align}

                Actually, QAP is regarded as an extremely difficult NP-hard problem. When the problem is large ($N > 30$), obtaining an exact solution is impossible. Therefore QAP is generally solved by approximating methods based on meta-heuristics~\cite{loiola2007survey}. Results of extensive experiments~\cite{drezner2008extensive} show that a hybrid algorithm of Genetic Algorithm and Tabu Search~\cite{drezner2003new} is effective. In this paper, we solve it using an algorithm based on them, but use Hill Climbing instead of Tabu Search because of computational efficiency.

                Our optimization procedure is shown in Algorithm~\ref{alg:optimize}. First, we generate random $N_P$ solutions, optimize each by Hill Climbing, and append to population $P$. Second, we select two solutions $p_1$, $p_2$ in $P$ randomly, generate a child $c$ from $p_1$, $p_2$, and optimize $c$ by Hill Climbing. If $c$ is better than the worst solution $p_w$ in $P$, then we append $c$ to $P$ and remove one solution. We exclude $p_w$ with a probability of $1-p$, or one of the most similar pairs in $P$ otherwise. We repeat this procedure until the best and worst solution become almost identical.
                
                We generate a random solution by assigning $N$ elements randomly to $N$ places without overlapping. At Hill Climbing, we modify a solution by exchanging the assignment of two elements, which improves the objective value the best. We generate a child and measure the similarity of two solutions in the manner described by Drezner et al.~\cite{drezner2003new}. We set $N_P=100$ and $p=0.2$.

                \begin{algorithm}[t]
                  \caption{Optimization of Equation~\ref{eq:cost}.}
                  \begin{algorithmic}
                    \REQUIRE{BoVW (to generate initial solutions), $C^{a}, C^{p}, \lambda$ (to evaluate objective values)}
                    \ENSURE{optimal arrangement $x^*$}
                    \STATE{population $P$ $\leftarrow$ randomly generated $N_P$ solutions}
                    \WHILE{$ \max (P)  \neq \min (P) $}
                      \STATE{$p_1, p_2$ $\leftarrow$ randomly selected pair in $P$}
                      \STATE{$c$ $\leftarrow$ a new child generated from $p_1, p_2$}
                      \STATE{optimize $c$ by Hill Climbing}
                    \IF{$c < \max(P)$ }
                      \STATE{append $c$ to $P$}
                      \IF{ $\mbox{rand}(0,1) < p $}
                        \STATE{remove one of the most similar pair in $P$}
                      \ELSE
                        \STATE{remove $\argmax (P)$}
                      \ENDIF
                      \ENDIF
                    \ENDWHILE
                    \RETURN{$\argmin (P)$}
                  \end{algorithmic}
                  \label{alg:optimize}
                \end{algorithm}

        \begin{figure*}[t]
          \begin{center}
            \raisebox{5mm}{\makebox[28mm][l]{\small{(a) Original image}}}
            \includegraphics[width=13.4mm,bb=0 0 128 128]{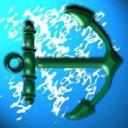}
            \includegraphics[width=13.4mm,bb=0 0 128 128]{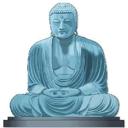}
            \includegraphics[width=13.4mm,bb=0 0 128 128]{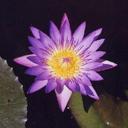}
            \includegraphics[width=13.4mm,bb=0 0 128 128]{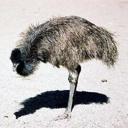}
            \includegraphics[width=13.4mm,bb=0 0 128 128]{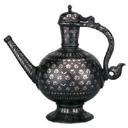}
            \includegraphics[width=13.4mm,bb=0 0 128 128]{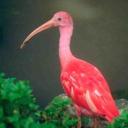}
            \includegraphics[width=13.4mm,bb=0 0 128 128]{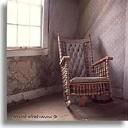}
            \includegraphics[width=13.4mm,bb=0 0 128 128]{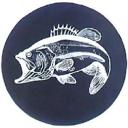}
            \includegraphics[width=13.4mm,bb=0 0 128 128]{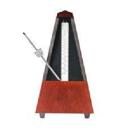}
            \includegraphics[width=13.4mm,bb=0 0 128 128]{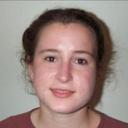} \\
            \raisebox{5mm}{\makebox[28mm][l]{\small{(b) Our method}}}
            \includegraphics[width=13.4mm,bb=0 0 128 128]{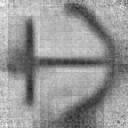}
            \includegraphics[width=13.4mm,bb=0 0 128 128]{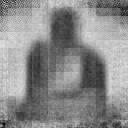}
            \includegraphics[width=13.4mm,bb=0 0 128 128]{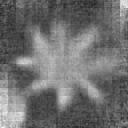}
            \includegraphics[width=13.4mm,bb=0 0 128 128]{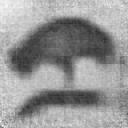}
            \includegraphics[width=13.4mm,bb=0 0 128 128]{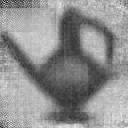}
            \includegraphics[width=13.4mm,bb=0 0 128 128]{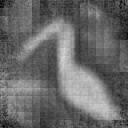}
            \includegraphics[width=13.4mm,bb=0 0 128 128]{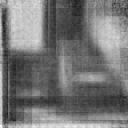}
            \includegraphics[width=13.4mm,bb=0 0 128 128]{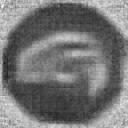}
            \includegraphics[width=13.4mm,bb=0 0 128 128]{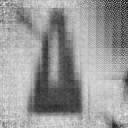}
            \includegraphics[width=13.4mm,bb=0 0 128 128]{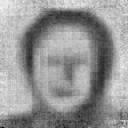} \\
            \raisebox{5mm}{\makebox[28mm][l]{\small{(c) HOGgles}}}
            \includegraphics[width=13.4mm,bb=0 0 128 128]{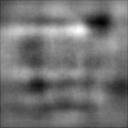}
            \includegraphics[width=13.4mm,bb=0 0 128 128]{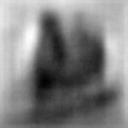}
            \includegraphics[width=13.4mm,bb=0 0 128 128]{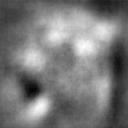}
            \includegraphics[width=13.4mm,bb=0 0 128 128]{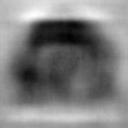}
            \includegraphics[width=13.4mm,bb=0 0 128 128]{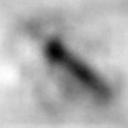}
            \includegraphics[width=13.4mm,bb=0 0 128 128]{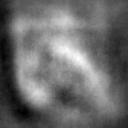}
            \includegraphics[width=13.4mm,bb=0 0 128 128]{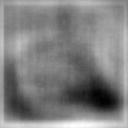}
            \includegraphics[width=13.4mm,bb=0 0 128 128]{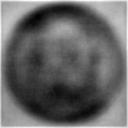}
            \includegraphics[width=13.4mm,bb=0 0 128 128]{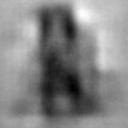}
            \includegraphics[width=13.4mm,bb=0 0 128 128]{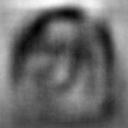} \\
            \raisebox{5mm}{\makebox[28mm][l]{\small{(d) Image retrieval}}}
            \includegraphics[width=13.4mm,bb=0 0 128 128]{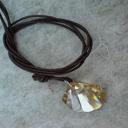}
            \includegraphics[width=13.4mm,bb=0 0 128 128]{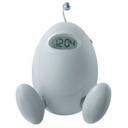}
            \includegraphics[width=13.4mm,bb=0 0 128 128]{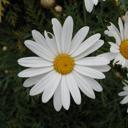}
            \includegraphics[width=13.4mm,bb=0 0 128 128]{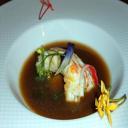}
            \includegraphics[width=13.4mm,bb=0 0 128 128]{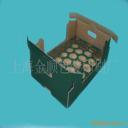}
            \includegraphics[width=13.4mm,bb=0 0 128 128]{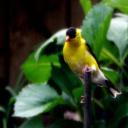}
            \includegraphics[width=13.4mm,bb=0 0 128 128]{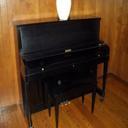}
            \includegraphics[width=13.4mm,bb=0 0 128 128]{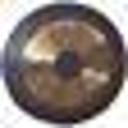}
            \includegraphics[width=13.4mm,bb=0 0 128 128]{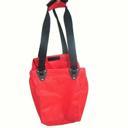}
            \includegraphics[width=13.4mm,bb=0 0 128 128]{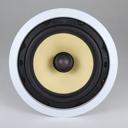}
          \end{center}
          \caption{Examples of images obtained from BoVW. (a) Original images to be reconstructed. (b) Images reconstructed using our method. (c) Images reconstructed by HOGgles~\cite{vondrick2012inverting}. (d) Nearest images retrieved from one million images. }
          \label{fig:reconstruced_from_bovw}
        \end{figure*}

        \subsection{Relation to Jigsaw Solvers}
            \label{sec:method_jigsaw_solver}

            As described in Section~\ref{sec:method_analogy_with_jigsaw}, our main problem is similar to solving jigsaw puzzles. However, automatic jigsaw solvers have been proposed in recent years. We show the relation between our method and such methods.

            The objective function of Cho et al.~\cite{cho2010probabilistic} includes global and local naturalness terms that are almost identical to ours. They solved it by Belief Propagation with approximation. Their method is not so accurate. Therefore to achieve fine results using their method, correct positions of some pieces must be given. Pomeranz et al.~\cite{pomeranz2011fully} considered only local naturalness, and optimized it by repeating greedy assignments and retrial of unacceptable regions. Gallagher et al.~\cite{gallagher2012jigsaw} assumed that the orientations of the pieces are also unknown in addition to their positions. They also considered only local naturalness. They proposed sophisticated method to evaluate whether two pieces are connected or not, and optimized the whole assignment by solving the minimum spanning tree. Sholomon et al.~\cite{sholomon2013genetic} proposed an efficient genetic algorithm and solved puzzles of more than $20,000$ pieces, which had been thought to be impossible.

            Although all of them consider one kind of Local Adjacent Cost, in jigsaw puzzle settings, $C^{a}_{ijkl}$ is a nonzero value only if $k$ is a $4$-neighbor of $l$. This feature reduces the structure of the graph of elements, making them much simpler; most of them use it. The Global Location Cost is included only in the model proposed by Cho et al.~\cite{cho2010probabilistic} and not in the others~\cite{pomeranz2011fully, gallagher2012jigsaw, sholomon2013genetic}. For these reasons, optimization of jigsaw puzzles is less complicated. Large problems are solvable efficiently.

        \subsection{Image Generation}
            \label{sec:method_generation}

            Once the spatial arrangement of visual words is estimated, the remaining task is to generate images from the set of visual words and their geometric information. This is accomplished by converting each visual word to an image patch and transforming and blending them into one image canvas.
            
            Several methods exist to convert a visual word (a local descriptor) into an image patch~\cite{weinzaepfel2011reconstructing, d2012beyond, vondrick2012inverting}. Among them, a method by Vondrick et al.\cite{vondrick2012inverting} performs well. It is applicable to descriptors of arbitrary kinds. We use it to image patch generation.
            
            Weinzaepfel et al.~\cite{weinzaepfel2011reconstructing} used Poisson Blending~\cite{perez2003poisson} to blend image patches. Because they used keypoint based sampling, the image patch sizes differ. They first blended larger patches to draw the overall structure; then they blended smaller ones to draw small details. As described herein, because we adopted dense sampling, the image patch sizes are equal, and it is desirable that generated images not to be dependent on the order of the patches to draw. Therefore, we simply use additive synthesis.

        \subsection{Evaluation Metrics}
            \label{sec:method_evaluation}
            
            The objective is to generate an image resembling the original image. Therefore, for evaluation, one must measure the similarity of the two images, which is an ultimate goal of computer vision and which has not been accomplished adequately. Various image features have been proposed to compute image similarity. However, to avoid arbitrariness of selecting features for evaluation, we do not use image features.
            
            The simplest option is to use the normalized cross correlation of pixel values of two images as used by Vondrick et al.\cite{vondrick2012inverting}. We designate it as {\it XCORR}. However, this metric is extremely sensitive to a small shift of image contents. Therefore, we transform one image slightly, compute the normalized cross correlation for each transform, and take their maximum. We designate {\it XCORR4} for a maximum shift of $\pm 4$ pixels and {\it XCORR8} for $\pm 8$ pixels.
            
            Normalized cross correlation does not always match human evaluations, as reported in~\cite{vondrick2012inverting}. Therefore, to compare different methods, we also asked one hundred people to select which image from different methods is the most similar to the original image on CrowdFlower\footnote{\url{http://www.crowdflower.com/}}. We designate it {\it HUMAN}. For example, if 60 people judged an image by method A as the most similar to the original and 40 people selected an image by method B, HUMAN is $\{0.60, 0.40\}$.

            The main point of our method is the estimation of the original layout of visual words. Its evaluation is rather simple. Cho et al.~\cite{cho2010probabilistic} proposed metrics to evaluate the correctness of the solved jigsaw puzzles. We use two of them which are used repeatedly in later papers. One is {\it Direct Comparison (DC)}, which is the portion of pieces that are placed in the correct positions. The other is {\it Neighbor Comparison (NC)}, which is the portion of pairs of pieces in $4$-neighbor that are correct neighbors.

    \section{Experiments in Image Reconstruction}
        \label{chap:experiment}

            \begin{figure*}[t]
              \begin{center}
                \raisebox{5mm}{\makebox[28mm][l]{\small{(a) No quantization}}}
                \includegraphics[width=13.4mm,bb=0 0 128 128]{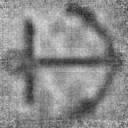}
                \includegraphics[width=13.4mm,bb=0 0 128 128]{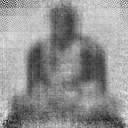}
                \includegraphics[width=13.4mm,bb=0 0 128 128]{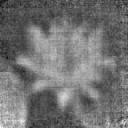}
                \includegraphics[width=13.4mm,bb=0 0 128 128]{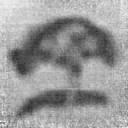}
                \includegraphics[width=13.4mm,bb=0 0 128 128]{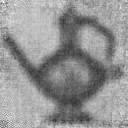}
                \includegraphics[width=13.4mm,bb=0 0 128 128]{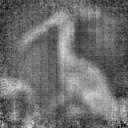}
                \includegraphics[width=13.4mm,bb=0 0 128 128]{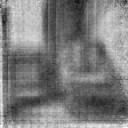}
                \includegraphics[width=13.4mm,bb=0 0 128 128]{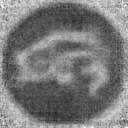}
                \includegraphics[width=13.4mm,bb=0 0 128 128]{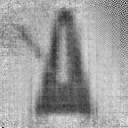}
                \includegraphics[width=13.4mm,bb=0 0 128 128]{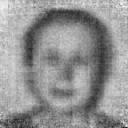} \\
                \raisebox{5mm}{\makebox[28mm][l]{\small{(b) $k=8192$}}}
                \includegraphics[width=13.4mm,bb=0 0 128 128]{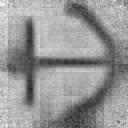}
                \includegraphics[width=13.4mm,bb=0 0 128 128]{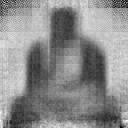}
                \includegraphics[width=13.4mm,bb=0 0 128 128]{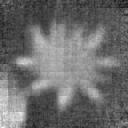}
                \includegraphics[width=13.4mm,bb=0 0 128 128]{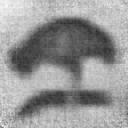}
                \includegraphics[width=13.4mm,bb=0 0 128 128]{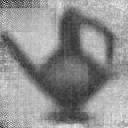}
                \includegraphics[width=13.4mm,bb=0 0 128 128]{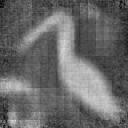}
                \includegraphics[width=13.4mm,bb=0 0 128 128]{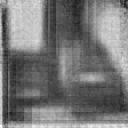}
                \includegraphics[width=13.4mm,bb=0 0 128 128]{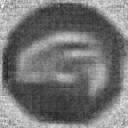}
                \includegraphics[width=13.4mm,bb=0 0 128 128]{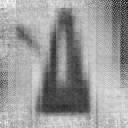}
                \includegraphics[width=13.4mm,bb=0 0 128 128]{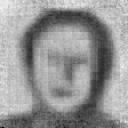} \\
                \raisebox{5mm}{\makebox[28mm][l]{\small{(c) $k=128$}}}
                \includegraphics[width=13.4mm,bb=0 0 128 128]{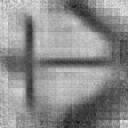}
                \includegraphics[width=13.4mm,bb=0 0 128 128]{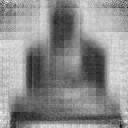}
                \includegraphics[width=13.4mm,bb=0 0 128 128]{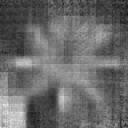}
                \includegraphics[width=13.4mm,bb=0 0 128 128]{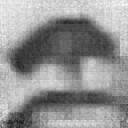}
                \includegraphics[width=13.4mm,bb=0 0 128 128]{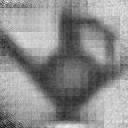}
                \includegraphics[width=13.4mm,bb=0 0 128 128]{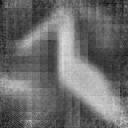}
                \includegraphics[width=13.4mm,bb=0 0 128 128]{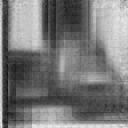}
                \includegraphics[width=13.4mm,bb=0 0 128 128]{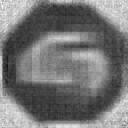}
                \includegraphics[width=13.4mm,bb=0 0 128 128]{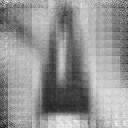}
                \includegraphics[width=13.4mm,bb=0 0 128 128]{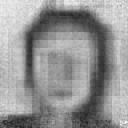} \\
              \end{center}
              \caption{Examples of images reconstructed from local descriptors and their geometric information. $k$ is the visual word dictionary size. Well identifiable images are reconstructed even if the descriptors are quantized. }
              \label{fig:reconstruced_aligned}
            \end{figure*}

            \begin{figure*}
              \begin{center}
                \raisebox{5mm}{\makebox[28mm][l]{\small{(a) Best results}}}
                \includegraphics[width=13.4mm,bb=0 0 128 128]{./data/original/010.jpg}
                \includegraphics[width=13.4mm,bb=0 0 128 128]{./data/proposed/SIFT_k_8192_lambda_08/010.jpg}
                \includegraphics[width=13.4mm,bb=0 0 128 128]{./data/original/040.jpg}
                \includegraphics[width=13.4mm,bb=0 0 128 128]{./data/proposed/SIFT_k_8192_lambda_08/040.jpg}
                \includegraphics[width=13.4mm,bb=0 0 128 128]{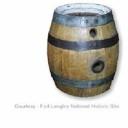}
                \includegraphics[width=13.4mm,bb=0 0 128 128]{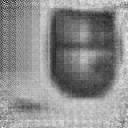}
                \includegraphics[width=13.4mm,bb=0 0 128 128]{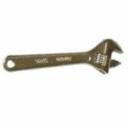}
                \includegraphics[width=13.4mm,bb=0 0 128 128]{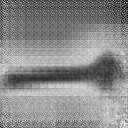}
                \includegraphics[width=13.4mm,bb=0 0 128 128]{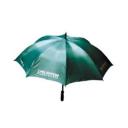}
                \includegraphics[width=13.4mm,bb=0 0 128 128]{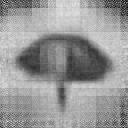} \\
                \raisebox{5mm}{\makebox[28mm][l]{\small{(b) Worst results}}}
                \includegraphics[width=13.4mm,bb=0 0 128 128]{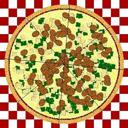}
                \includegraphics[width=13.4mm,bb=0 0 128 128]{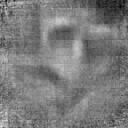}
                \includegraphics[width=13.4mm,bb=0 0 128 128]{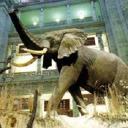}
                \includegraphics[width=13.4mm,bb=0 0 128 128]{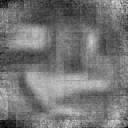}
                \includegraphics[width=13.4mm,bb=0 0 128 128]{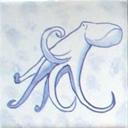}
                \includegraphics[width=13.4mm,bb=0 0 128 128]{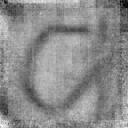}
                \includegraphics[width=13.4mm,bb=0 0 128 128]{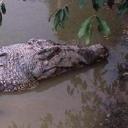}
                \includegraphics[width=13.4mm,bb=0 0 128 128]{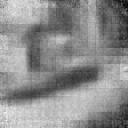}
                \includegraphics[width=13.4mm,bb=0 0 128 128]{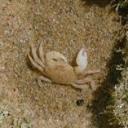}
                \includegraphics[width=13.4mm,bb=0 0 128 128]{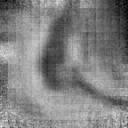} \\
              \end{center}
              \caption{Best and the worst results. The top left signifies the best. The bottom left represents the worst. Images are sorted by their neighbor comparison score. These results demonstrate that the arrangement accuracy is dependent on the complexity of the image.}
              \label{fig:experiment_best_worst}
            \end{figure*}

         In this section, we present an evaluation of our proposed method through experiments in reconstruction. First, we demonstrate that our method can reconstruct images from BoVW. Then we compare ours with other approaches. Thereafter, we examine the effect of the weighting parameter that balances global and local naturalness, the size of the visual word dictionary, and the mode of optimization. We also show reconstructed images derived from local descriptors of various kinds.

        We built a dataset to evaluate the reconstruction performance. We randomly extracted one image per class from Caltech $101$ Dataset~\cite{fei2007learning} and appended them to our dataset. Caltech $101$ Dataset has $101$ object classes. Therefore, our dataset includes $101$ images of $101$ kinds of objects. Some images in our dataset are shown in Fig.~\ref{fig:reconstruced_from_bovw}(a). All images in the dataset are shown in Fig.~\ref{fig:appendix_all_images}.

        The experiment settings are as follows unless otherwise noted. We set the vocabulary size of the visual word dictionary (which is identical to the dimension of a BoVW) to $8192$, the size of an image patch for extracting a local descriptor to $32 \times 32$ pixels, and the extraction step of descriptors to $8$ pixels. We resized all the images used in the experiments to $128 \times 128$ pixels. We used SIFT descriptors~\cite{lowe2004distinctive} as local descriptors. We used one million images extracted randomly from ILSVRC 2012 image classification dataset\footnote{\url{http://www.image-net.org/challenges/LSVRC/2012/}} to construct Global Position Cost and Local Adjacency Cost. In this setting, a BoVW includes $13 \times 13$ visual words. Three-fourths of a visual word overlapping with the next visual word.

        Codes of all experiments are available from the author's website\footnote{\url{http://hiroharu-kato.com/}}.
        
        \subsection{Image Reconstruction from Visual Words and Their Geometric Information}
            \label{sec:experiment_aligned}

            Our proposed method can be decomposed into 1) estimation of optimal spatial arrangement of visual words, and 2) generation of an image from visual words and their position information. To elucidate the capability of the method, in this section, we presume that the layout of visual words is given. We examine the latter procedure. Additionally, we explore the effect of the quantization error which occurs when local descriptors are assigned to visual words at the coding step.

            Fig.~\ref{fig:reconstruced_aligned} shows reconstructed images from the local descriptor and their geometric information. Local descriptors in Fig.~\ref{fig:reconstruced_aligned}(a) are not quantized, and those in Figs.~\ref{fig:reconstruced_aligned}(b) and \ref{fig:reconstruced_aligned}(c) are quantized to $8192$ or $128$ visual words in the dictionary. Although they are less clear than the original images in Fig.~\ref{fig:reconstruced_from_bovw}(a), their image contents are still identifiable. If the dictionary becomes smaller, then the quantization error in coding step becomes higher; the resulting images become more blurred. However they are still identifiable. They become grayscale images because SIFT descriptors do not capture color information.

        \subsection{Image Reconstruction from BoVW}
            \label{sec:experiment_bovw}
            
            This section describes confirmation that our method can reconstruct images from BoVW. Additionally, we compare ours with the following two methods. 
            \begin{enumerate}
                \item A feature visualization method called HOGgles was proposed by Vondrick et al.~\cite{vondrick2012inverting}. Although it is applicable to arbitrary image features in principle, they reported the results of HOG and HOG-based features only. 
                \item Similar image retrieval using BoVW. BoVW demonstrates superior performance for image search. Therefore, the results of retrieval by BoVW might help us to ascertain what BoVW captures. This experiment examines the nearest image measured by Euclidean distance. We used one million images from ILSVRC 2012 image classification dataset as the source of image retrieval.
            \end{enumerate}

            \begin{table}
              \caption{Quantitative evaluation of the proposed method, HOGgles~\cite{vondrick2012inverting}, and image retrieval (IR).  Details of metrics are described in Section~\ref{sec:method_evaluation}. Scores are averaged over $101$ images in the reconstruction dataset.}
              \begin{center}
                \begin{tabular}{|l||c|c|c||c|}
                  \hline
                          & XCORR              & XCORR4           & XCORR8           & HUMAN            \\
                  \hline\hline
                  Ours    & $ 0.408 $          & $\mathbf{0.489}$ & $\mathbf{0.542}$ & $\mathbf{0.874}$ \\
                  HOGgles & $ \mathbf{0.409} $ & $0.425$          & $0.434$          & $0.065$          \\
                  IR      & $ 0.233 $          & $0.286$          & $0.333$          & $0.061$          \\
                  \hline
                \end{tabular}
              \end{center}
              \label{table:experiment_quantitative1}
            \end{table}

            \begin{figure*}
                \begin{center}
                    \raisebox{5mm}{\makebox[28mm][l]{\small{(a) $\lambda=0.0$}}}
                    \includegraphics[width=13.4mm,bb=0 0 128 128]{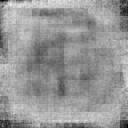}
                    \includegraphics[width=13.4mm,bb=0 0 128 128]{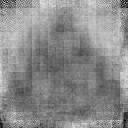}
                    \includegraphics[width=13.4mm,bb=0 0 128 128]{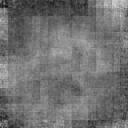}
                    \includegraphics[width=13.4mm,bb=0 0 128 128]{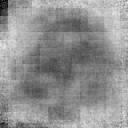}
                    \includegraphics[width=13.4mm,bb=0 0 128 128]{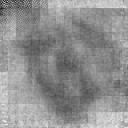}
                    \includegraphics[width=13.4mm,bb=0 0 128 128]{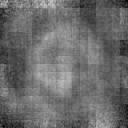}
                    \includegraphics[width=13.4mm,bb=0 0 128 128]{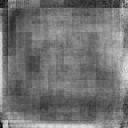}
                    \includegraphics[width=13.4mm,bb=0 0 128 128]{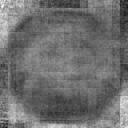}
                    \includegraphics[width=13.4mm,bb=0 0 128 128]{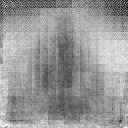}
                    \includegraphics[width=13.4mm,bb=0 0 128 128]{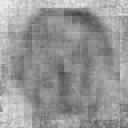} \\
                    \raisebox{5mm}{\makebox[28mm][l]{\small{(b) $\lambda=0.2$}}}
                    \includegraphics[width=13.4mm,bb=0 0 128 128]{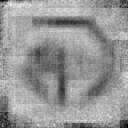}
                    \includegraphics[width=13.4mm,bb=0 0 128 128]{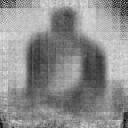}
                    \includegraphics[width=13.4mm,bb=0 0 128 128]{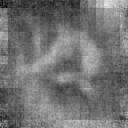}
                    \includegraphics[width=13.4mm,bb=0 0 128 128]{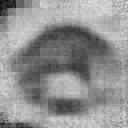}
                    \includegraphics[width=13.4mm,bb=0 0 128 128]{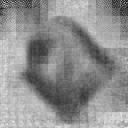}
                    \includegraphics[width=13.4mm,bb=0 0 128 128]{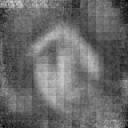}
                    \includegraphics[width=13.4mm,bb=0 0 128 128]{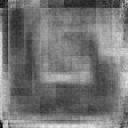}
                    \includegraphics[width=13.4mm,bb=0 0 128 128]{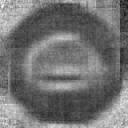}
                    \includegraphics[width=13.4mm,bb=0 0 128 128]{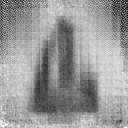}
                    \includegraphics[width=13.4mm,bb=0 0 128 128]{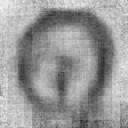} \\
                    \raisebox{5mm}{\makebox[28mm][l]{\small{(b) $\lambda=0.8$}}}
                    \includegraphics[width=13.4mm,bb=0 0 128 128]{./data/proposed/SIFT_k_8192_lambda_08/007.jpg}
                    \includegraphics[width=13.4mm,bb=0 0 128 128]{./data/proposed/SIFT_k_8192_lambda_08/016.jpg}
                    \includegraphics[width=13.4mm,bb=0 0 128 128]{./data/proposed/SIFT_k_8192_lambda_08/096.jpg}
                    \includegraphics[width=13.4mm,bb=0 0 128 128]{./data/proposed/SIFT_k_8192_lambda_08/038.jpg}
                    \includegraphics[width=13.4mm,bb=0 0 128 128]{./data/proposed/SIFT_k_8192_lambda_08/040.jpg}
                    \includegraphics[width=13.4mm,bb=0 0 128 128]{./data/proposed/SIFT_k_8192_lambda_08/052.jpg}
                    \includegraphics[width=13.4mm,bb=0 0 128 128]{./data/proposed/SIFT_k_8192_lambda_08/023.jpg}
                    \includegraphics[width=13.4mm,bb=0 0 128 128]{./data/proposed/SIFT_k_8192_lambda_08/010.jpg}
                    \includegraphics[width=13.4mm,bb=0 0 128 128]{./data/proposed/SIFT_k_8192_lambda_08/065.jpg}
                    \includegraphics[width=13.4mm,bb=0 0 128 128]{./data/proposed/SIFT_k_8192_lambda_08/002.jpg} \\
                    \raisebox{5mm}{\makebox[28mm][l]{\small{(c) $\lambda=1.0$}}}
                    \includegraphics[width=13.4mm,bb=0 0 128 128]{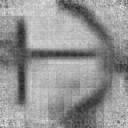}
                    \includegraphics[width=13.4mm,bb=0 0 128 128]{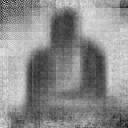}
                    \includegraphics[width=13.4mm,bb=0 0 128 128]{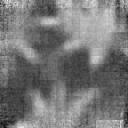}
                    \includegraphics[width=13.4mm,bb=0 0 128 128]{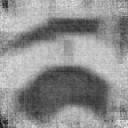}
                    \includegraphics[width=13.4mm,bb=0 0 128 128]{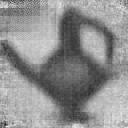}
                    \includegraphics[width=13.4mm,bb=0 0 128 128]{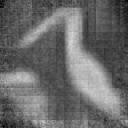}
                    \includegraphics[width=13.4mm,bb=0 0 128 128]{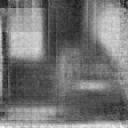}
                    \includegraphics[width=13.4mm,bb=0 0 128 128]{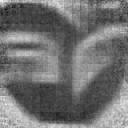}
                    \includegraphics[width=13.4mm,bb=0 0 128 128]{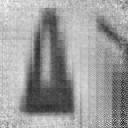}
                    \includegraphics[width=13.4mm,bb=0 0 128 128]{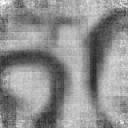} \\
                \end{center}
                \caption{Examples of reconstructed images. The parameter $\lambda$, which balances global and local naturalness of visual words, is varied. Local naturalness is ignored at $\lambda=0.0$; global naturalness is excluded at $\lambda=1.0$.}
                \label{fig:experiment_lambda}
            \end{figure*}

\begin{figure*}[p]
            \begin{center}
                \includegraphics[width=9.6mm,bb=0 0 128 128]{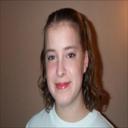}
                \includegraphics[width=9.6mm,bb=0 0 128 128]{./data/original/002.jpg}
                \includegraphics[width=9.6mm,bb=0 0 128 128]{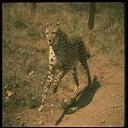}
                \includegraphics[width=9.6mm,bb=0 0 128 128]{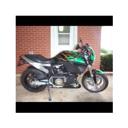}
                \includegraphics[width=9.6mm,bb=0 0 128 128]{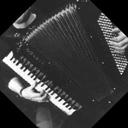}
                \includegraphics[width=9.6mm,bb=0 0 128 128]{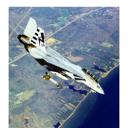}
                \includegraphics[width=9.6mm,bb=0 0 128 128]{./data/original/007.jpg}
                \includegraphics[width=9.6mm,bb=0 0 128 128]{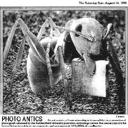}
                \includegraphics[width=9.6mm,bb=0 0 128 128]{./data/original/009.jpg}
                \includegraphics[width=9.6mm,bb=0 0 128 128]{./data/original/010.jpg}
                \includegraphics[width=9.6mm,bb=0 0 128 128]{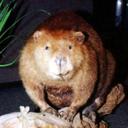}
                \includegraphics[width=9.6mm,bb=0 0 128 128]{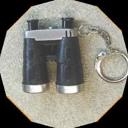}
                \includegraphics[width=9.6mm,bb=0 0 128 128]{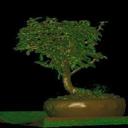}
                \includegraphics[width=9.6mm,bb=0 0 128 128]{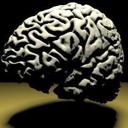}
                \includegraphics[width=9.6mm,bb=0 0 128 128]{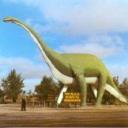}
                \includegraphics[width=9.6mm,bb=0 0 128 128]{./data/original/016.jpg}
                \includegraphics[width=9.6mm,bb=0 0 128 128]{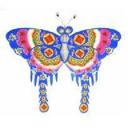} \\
                \includegraphics[width=9.6mm,bb=0 0 128 128]{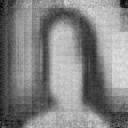}
                \includegraphics[width=9.6mm,bb=0 0 128 128]{./data/proposed/SIFT_k_8192_lambda_08/002.jpg}
                \includegraphics[width=9.6mm,bb=0 0 128 128]{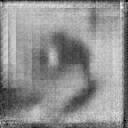}
                \includegraphics[width=9.6mm,bb=0 0 128 128]{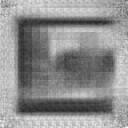}
                \includegraphics[width=9.6mm,bb=0 0 128 128]{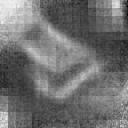}
                \includegraphics[width=9.6mm,bb=0 0 128 128]{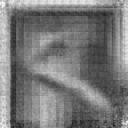}
                \includegraphics[width=9.6mm,bb=0 0 128 128]{./data/proposed/SIFT_k_8192_lambda_08/007.jpg}
                \includegraphics[width=9.6mm,bb=0 0 128 128]{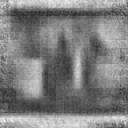}
                \includegraphics[width=9.6mm,bb=0 0 128 128]{./data/proposed/SIFT_k_8192_lambda_08/009.jpg}
                \includegraphics[width=9.6mm,bb=0 0 128 128]{./data/proposed/SIFT_k_8192_lambda_08/010.jpg}
                \includegraphics[width=9.6mm,bb=0 0 128 128]{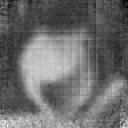}
                \includegraphics[width=9.6mm,bb=0 0 128 128]{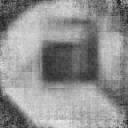}
                \includegraphics[width=9.6mm,bb=0 0 128 128]{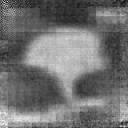}
                \includegraphics[width=9.6mm,bb=0 0 128 128]{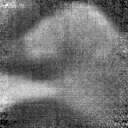}
                \includegraphics[width=9.6mm,bb=0 0 128 128]{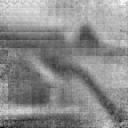}
                \includegraphics[width=9.6mm,bb=0 0 128 128]{./data/proposed/SIFT_k_8192_lambda_08/016.jpg}
                \includegraphics[width=9.6mm,bb=0 0 128 128]{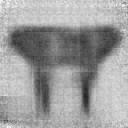} \\ \vspace{2mm}

                \includegraphics[width=9.6mm,bb=0 0 128 128]{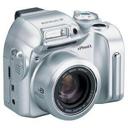}
                \includegraphics[width=9.6mm,bb=0 0 128 128]{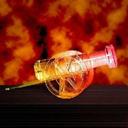}
                \includegraphics[width=9.6mm,bb=0 0 128 128]{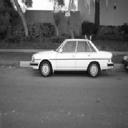}
                \includegraphics[width=9.6mm,bb=0 0 128 128]{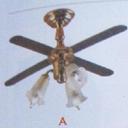}
                \includegraphics[width=9.6mm,bb=0 0 128 128]{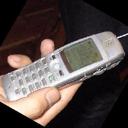}
                \includegraphics[width=9.6mm,bb=0 0 128 128]{./data/original/023.jpg}
                \includegraphics[width=9.6mm,bb=0 0 128 128]{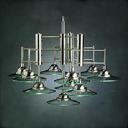}
                \includegraphics[width=9.6mm,bb=0 0 128 128]{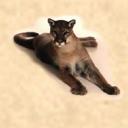}
                \includegraphics[width=9.6mm,bb=0 0 128 128]{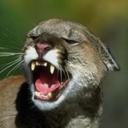}
                \includegraphics[width=9.6mm,bb=0 0 128 128]{./data/original/027.jpg}
                \includegraphics[width=9.6mm,bb=0 0 128 128]{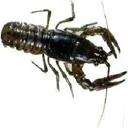}
                \includegraphics[width=9.6mm,bb=0 0 128 128]{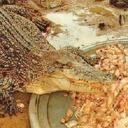}
                \includegraphics[width=9.6mm,bb=0 0 128 128]{./data/original/030.jpg}
                \includegraphics[width=9.6mm,bb=0 0 128 128]{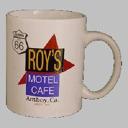}
                \includegraphics[width=9.6mm,bb=0 0 128 128]{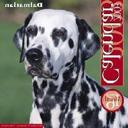}
                \includegraphics[width=9.6mm,bb=0 0 128 128]{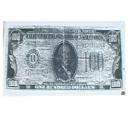}
                \includegraphics[width=9.6mm,bb=0 0 128 128]{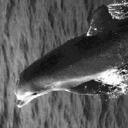} \\
                \includegraphics[width=9.6mm,bb=0 0 128 128]{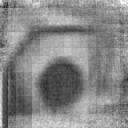}
                \includegraphics[width=9.6mm,bb=0 0 128 128]{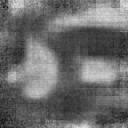}
                \includegraphics[width=9.6mm,bb=0 0 128 128]{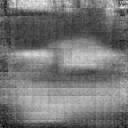}
                \includegraphics[width=9.6mm,bb=0 0 128 128]{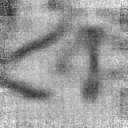}
                \includegraphics[width=9.6mm,bb=0 0 128 128]{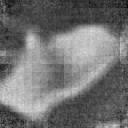}
                \includegraphics[width=9.6mm,bb=0 0 128 128]{./data/proposed/SIFT_k_8192_lambda_08/023.jpg}
                \includegraphics[width=9.6mm,bb=0 0 128 128]{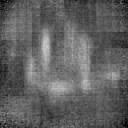}
                \includegraphics[width=9.6mm,bb=0 0 128 128]{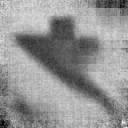}
                \includegraphics[width=9.6mm,bb=0 0 128 128]{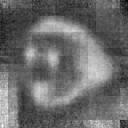}
                \includegraphics[width=9.6mm,bb=0 0 128 128]{./data/proposed/SIFT_k_8192_lambda_08/027.jpg}
                \includegraphics[width=9.6mm,bb=0 0 128 128]{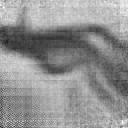}
                \includegraphics[width=9.6mm,bb=0 0 128 128]{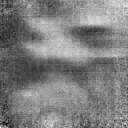}
                \includegraphics[width=9.6mm,bb=0 0 128 128]{./data/proposed/SIFT_k_8192_lambda_08/030.jpg}
                \includegraphics[width=9.6mm,bb=0 0 128 128]{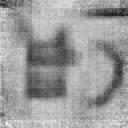}
                \includegraphics[width=9.6mm,bb=0 0 128 128]{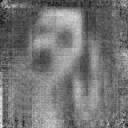}
                \includegraphics[width=9.6mm,bb=0 0 128 128]{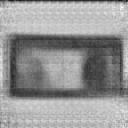}
                \includegraphics[width=9.6mm,bb=0 0 128 128]{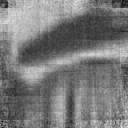} \\ \vspace{2mm}

                \includegraphics[width=9.6mm,bb=0 0 128 128]{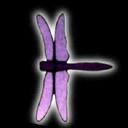}
                \includegraphics[width=9.6mm,bb=0 0 128 128]{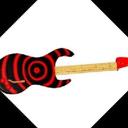}
                \includegraphics[width=9.6mm,bb=0 0 128 128]{./data/original/037.jpg}
                \includegraphics[width=9.6mm,bb=0 0 128 128]{./data/original/038.jpg}
                \includegraphics[width=9.6mm,bb=0 0 128 128]{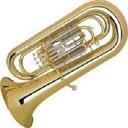}
                \includegraphics[width=9.6mm,bb=0 0 128 128]{./data/original/040.jpg}
                \includegraphics[width=9.6mm,bb=0 0 128 128]{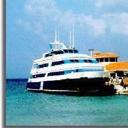}
                \includegraphics[width=9.6mm,bb=0 0 128 128]{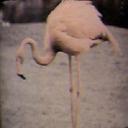}
                \includegraphics[width=9.6mm,bb=0 0 128 128]{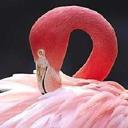}
                \includegraphics[width=9.6mm,bb=0 0 128 128]{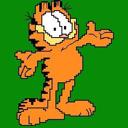}
                \includegraphics[width=9.6mm,bb=0 0 128 128]{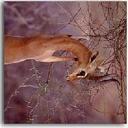}
                \includegraphics[width=9.6mm,bb=0 0 128 128]{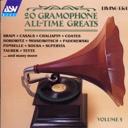}
                \includegraphics[width=9.6mm,bb=0 0 128 128]{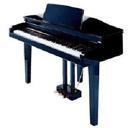}
                \includegraphics[width=9.6mm,bb=0 0 128 128]{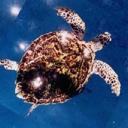}
                \includegraphics[width=9.6mm,bb=0 0 128 128]{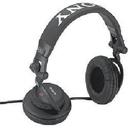}
                \includegraphics[width=9.6mm,bb=0 0 128 128]{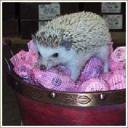}
                \includegraphics[width=9.6mm,bb=0 0 128 128]{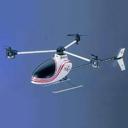} \\
                \includegraphics[width=9.6mm,bb=0 0 128 128]{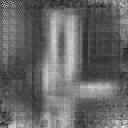}
                \includegraphics[width=9.6mm,bb=0 0 128 128]{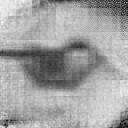}
                \includegraphics[width=9.6mm,bb=0 0 128 128]{./data/proposed/SIFT_k_8192_lambda_08/037.jpg}
                \includegraphics[width=9.6mm,bb=0 0 128 128]{./data/proposed/SIFT_k_8192_lambda_08/038.jpg}
                \includegraphics[width=9.6mm,bb=0 0 128 128]{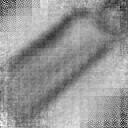}
                \includegraphics[width=9.6mm,bb=0 0 128 128]{./data/proposed/SIFT_k_8192_lambda_08/040.jpg}
                \includegraphics[width=9.6mm,bb=0 0 128 128]{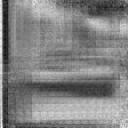}
                \includegraphics[width=9.6mm,bb=0 0 128 128]{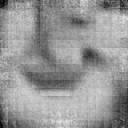}
                \includegraphics[width=9.6mm,bb=0 0 128 128]{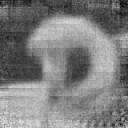}
                \includegraphics[width=9.6mm,bb=0 0 128 128]{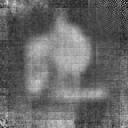}
                \includegraphics[width=9.6mm,bb=0 0 128 128]{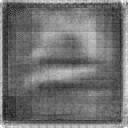}
                \includegraphics[width=9.6mm,bb=0 0 128 128]{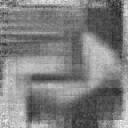}
                \includegraphics[width=9.6mm,bb=0 0 128 128]{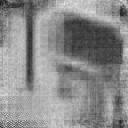}
                \includegraphics[width=9.6mm,bb=0 0 128 128]{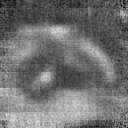}
                \includegraphics[width=9.6mm,bb=0 0 128 128]{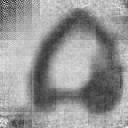}
                \includegraphics[width=9.6mm,bb=0 0 128 128]{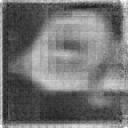}
                \includegraphics[width=9.6mm,bb=0 0 128 128]{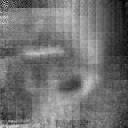} \\ \vspace{2mm}

                \includegraphics[width=9.6mm,bb=0 0 128 128]{./data/original/052.jpg}
                \includegraphics[width=9.6mm,bb=0 0 128 128]{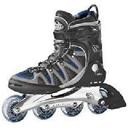}
                \includegraphics[width=9.6mm,bb=0 0 128 128]{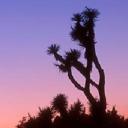}
                \includegraphics[width=9.6mm,bb=0 0 128 128]{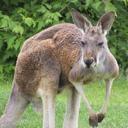}
                \includegraphics[width=9.6mm,bb=0 0 128 128]{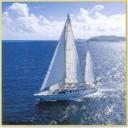}
                \includegraphics[width=9.6mm,bb=0 0 128 128]{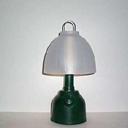}
                \includegraphics[width=9.6mm,bb=0 0 128 128]{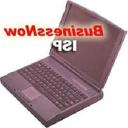}
                \includegraphics[width=9.6mm,bb=0 0 128 128]{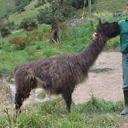}
                \includegraphics[width=9.6mm,bb=0 0 128 128]{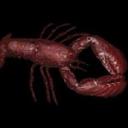}
                \includegraphics[width=9.6mm,bb=0 0 128 128]{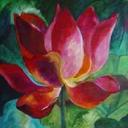}
                \includegraphics[width=9.6mm,bb=0 0 128 128]{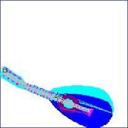}
                \includegraphics[width=9.6mm,bb=0 0 128 128]{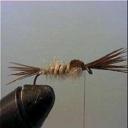}
                \includegraphics[width=9.6mm,bb=0 0 128 128]{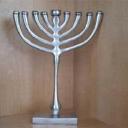}
                \includegraphics[width=9.6mm,bb=0 0 128 128]{./data/original/065.jpg}
                \includegraphics[width=9.6mm,bb=0 0 128 128]{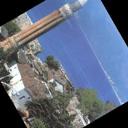}
                \includegraphics[width=9.6mm,bb=0 0 128 128]{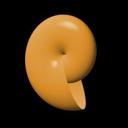}
                \includegraphics[width=9.6mm,bb=0 0 128 128]{./data/original/068.jpg}  \\
                \includegraphics[width=9.6mm,bb=0 0 128 128]{./data/proposed/SIFT_k_8192_lambda_08/052.jpg}
                \includegraphics[width=9.6mm,bb=0 0 128 128]{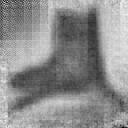}
                \includegraphics[width=9.6mm,bb=0 0 128 128]{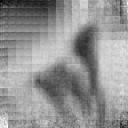}
                \includegraphics[width=9.6mm,bb=0 0 128 128]{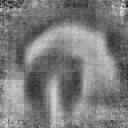}
                \includegraphics[width=9.6mm,bb=0 0 128 128]{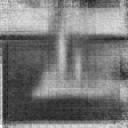}
                \includegraphics[width=9.6mm,bb=0 0 128 128]{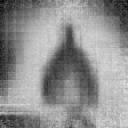}
                \includegraphics[width=9.6mm,bb=0 0 128 128]{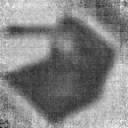}
                \includegraphics[width=9.6mm,bb=0 0 128 128]{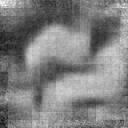}
                \includegraphics[width=9.6mm,bb=0 0 128 128]{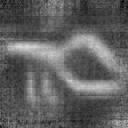}
                \includegraphics[width=9.6mm,bb=0 0 128 128]{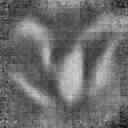}
                \includegraphics[width=9.6mm,bb=0 0 128 128]{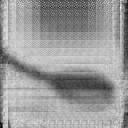}
                \includegraphics[width=9.6mm,bb=0 0 128 128]{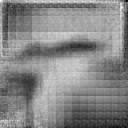}
                \includegraphics[width=9.6mm,bb=0 0 128 128]{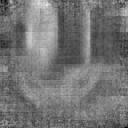}
                \includegraphics[width=9.6mm,bb=0 0 128 128]{./data/proposed/SIFT_k_8192_lambda_08/065.jpg}
                \includegraphics[width=9.6mm,bb=0 0 128 128]{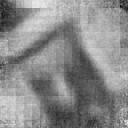}
                \includegraphics[width=9.6mm,bb=0 0 128 128]{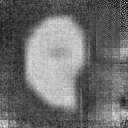}
                \includegraphics[width=9.6mm,bb=0 0 128 128]{./data/proposed/SIFT_k_8192_lambda_08/068.jpg} \\ \vspace{2mm}

                \includegraphics[width=9.6mm,bb=0 0 128 128]{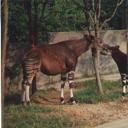}
                \includegraphics[width=9.6mm,bb=0 0 128 128]{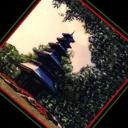}
                \includegraphics[width=9.6mm,bb=0 0 128 128]{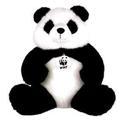}
                \includegraphics[width=9.6mm,bb=0 0 128 128]{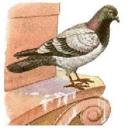}
                \includegraphics[width=9.6mm,bb=0 0 128 128]{./data/original/073.jpg}
                \includegraphics[width=9.6mm,bb=0 0 128 128]{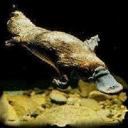}
                \includegraphics[width=9.6mm,bb=0 0 128 128]{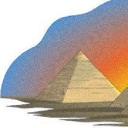}
                \includegraphics[width=9.6mm,bb=0 0 128 128]{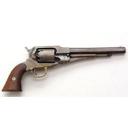}
                \includegraphics[width=9.6mm,bb=0 0 128 128]{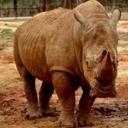}
                \includegraphics[width=9.6mm,bb=0 0 128 128]{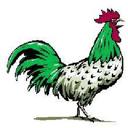}
                \includegraphics[width=9.6mm,bb=0 0 128 128]{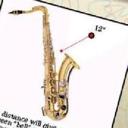}
                \includegraphics[width=9.6mm,bb=0 0 128 128]{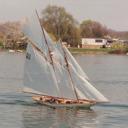}
                \includegraphics[width=9.6mm,bb=0 0 128 128]{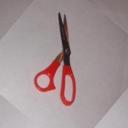}
                \includegraphics[width=9.6mm,bb=0 0 128 128]{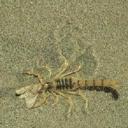}
                \includegraphics[width=9.6mm,bb=0 0 128 128]{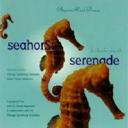}
                \includegraphics[width=9.6mm,bb=0 0 128 128]{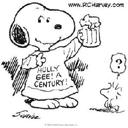}
                \includegraphics[width=9.6mm,bb=0 0 128 128]{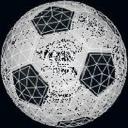} \\
                \includegraphics[width=9.6mm,bb=0 0 128 128]{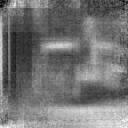}
                \includegraphics[width=9.6mm,bb=0 0 128 128]{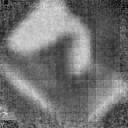}
                \includegraphics[width=9.6mm,bb=0 0 128 128]{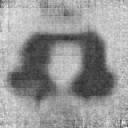}
                \includegraphics[width=9.6mm,bb=0 0 128 128]{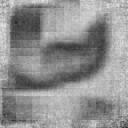}
                \includegraphics[width=9.6mm,bb=0 0 128 128]{./data/proposed/SIFT_k_8192_lambda_08/073.jpg}
                \includegraphics[width=9.6mm,bb=0 0 128 128]{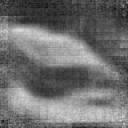}
                \includegraphics[width=9.6mm,bb=0 0 128 128]{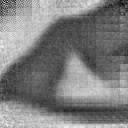}
                \includegraphics[width=9.6mm,bb=0 0 128 128]{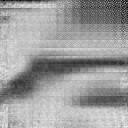}
                \includegraphics[width=9.6mm,bb=0 0 128 128]{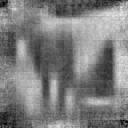}
                \includegraphics[width=9.6mm,bb=0 0 128 128]{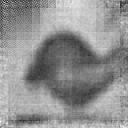}
                \includegraphics[width=9.6mm,bb=0 0 128 128]{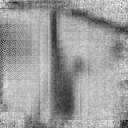}
                \includegraphics[width=9.6mm,bb=0 0 128 128]{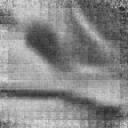}
                \includegraphics[width=9.6mm,bb=0 0 128 128]{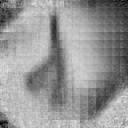}
                \includegraphics[width=9.6mm,bb=0 0 128 128]{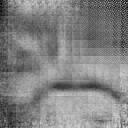}
                \includegraphics[width=9.6mm,bb=0 0 128 128]{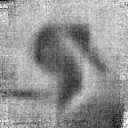}
                \includegraphics[width=9.6mm,bb=0 0 128 128]{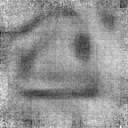}
                \includegraphics[width=9.6mm,bb=0 0 128 128]{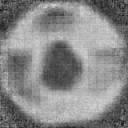} \\ \vspace{2mm}

                \includegraphics[width=9.6mm,bb=0 0 128 128]{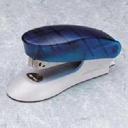}
                \includegraphics[width=9.6mm,bb=0 0 128 128]{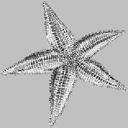}
                \includegraphics[width=9.6mm,bb=0 0 128 128]{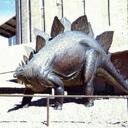}
                \includegraphics[width=9.6mm,bb=0 0 128 128]{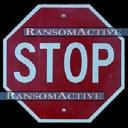}
                \includegraphics[width=9.6mm,bb=0 0 128 128]{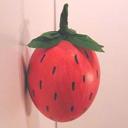}
                \includegraphics[width=9.6mm,bb=0 0 128 128]{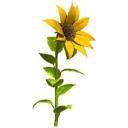}
                \includegraphics[width=9.6mm,bb=0 0 128 128]{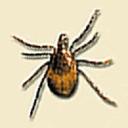}
                \includegraphics[width=9.6mm,bb=0 0 128 128]{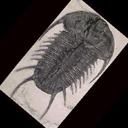}
                \includegraphics[width=9.6mm,bb=0 0 128 128]{./data/original/094.jpg}
                \includegraphics[width=9.6mm,bb=0 0 128 128]{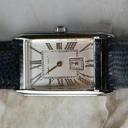}
                \includegraphics[width=9.6mm,bb=0 0 128 128]{./data/original/096.jpg}
                \includegraphics[width=9.6mm,bb=0 0 128 128]{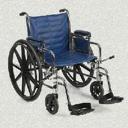}
                \includegraphics[width=9.6mm,bb=0 0 128 128]{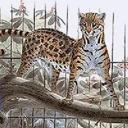}
                \includegraphics[width=9.6mm,bb=0 0 128 128]{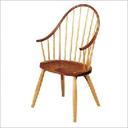}
                \includegraphics[width=9.6mm,bb=0 0 128 128]{./data/original/100.jpg}
                \includegraphics[width=9.6mm,bb=0 0 128 128]{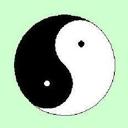}
                \includegraphics[width=9.6mm,bb=0 0 128 128]{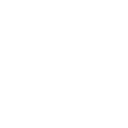} \\
                \includegraphics[width=9.6mm,bb=0 0 128 128]{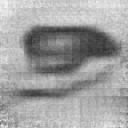}
                \includegraphics[width=9.6mm,bb=0 0 128 128]{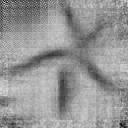}
                \includegraphics[width=9.6mm,bb=0 0 128 128]{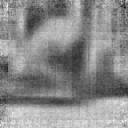}
                \includegraphics[width=9.6mm,bb=0 0 128 128]{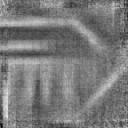}
                \includegraphics[width=9.6mm,bb=0 0 128 128]{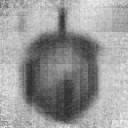}
                \includegraphics[width=9.6mm,bb=0 0 128 128]{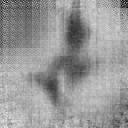}
                \includegraphics[width=9.6mm,bb=0 0 128 128]{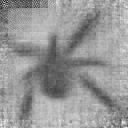}
                \includegraphics[width=9.6mm,bb=0 0 128 128]{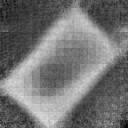}
                \includegraphics[width=9.6mm,bb=0 0 128 128]{./data/proposed/SIFT_k_8192_lambda_08/094.jpg}
                \includegraphics[width=9.6mm,bb=0 0 128 128]{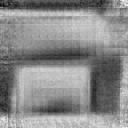}
                \includegraphics[width=9.6mm,bb=0 0 128 128]{./data/proposed/SIFT_k_8192_lambda_08/096.jpg}
                \includegraphics[width=9.6mm,bb=0 0 128 128]{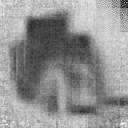}
                \includegraphics[width=9.6mm,bb=0 0 128 128]{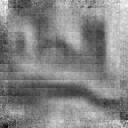}
                \includegraphics[width=9.6mm,bb=0 0 128 128]{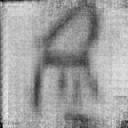}
                \includegraphics[width=9.6mm,bb=0 0 128 128]{./data/proposed/SIFT_k_8192_lambda_08/100.jpg}
                \includegraphics[width=9.6mm,bb=0 0 128 128]{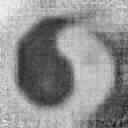}
                \includegraphics[width=9.6mm,bb=0 0 128 128]{./data/blank.jpg} \\ \vspace{2mm}
            \end{center}
            \caption{All images in our dataset and images reconstructed using  our method.}
            \label{fig:appendix_all_images}
        \end{figure*}

            Figs.~\ref{fig:reconstruced_from_bovw}(b)--\ref{fig:reconstruced_from_bovw}(d) present results obtained using each method. Some objects are visible in the images using our methods. They closely resemble the images in Fig.~\ref{fig:reconstruced_aligned}, which indicates that the positions of visual words are estimated properly. Although rough shapes of the images by HOGgles are the same as the original images, they are severely blurred and difficult to identify. Most images obtained by image retrieval are semantically different from the original images. Therefore, reading their original contents is difficult. All images reconstructed using our method are shown Fig.~\ref{fig:appendix_all_images}.

            It has been said that loss of spatial information of visual words is an important shortcoming of BoVW. There is plenty of work that has been done to compensate for it by embedding spatial information into BoVW~\cite{lazebnik2006beyond, harada2012graphical}. However, the results in Fig.~\ref{fig:reconstruced_from_bovw}(b) show that much spatial information is potentially retained in BoVW. Developing a method to embed spatial information into features by referring to reconstructed images might be an interesting approach.

            Table~\ref{table:experiment_quantitative1} presents quantitative evaluation of three methods. Details of the metrics are described in Section~\ref{sec:method_evaluation}. Our method outperforms other methods in view of XCORR$4$ and XCORR$8$. Ours is slightly inferior to HOGgles for XCORR because ours sometimes produces images of spatially shifted objects. Although the difference between our approach and others seems slight by seeing XCORR$n$, human evaluations strongly indicate the superiority of our method. Indeed, $87.4\%$ of images by our method are judged as most similar to the original.

            Fig.~\ref{fig:experiment_best_worst} shows five of the best and the worst results as evaluated by Neighbor Comparison. From this result, we can understand that our method performs better in reconstructing images of simple objects without background textures. However, it is weak in estimating objects that have complicated shapes or textures. This tendency indicates that local descriptors used here cannot acquire adequate information of complicated shapes. To improve representations for such images, we must use other kinds of descriptors or increase the size of the visual words dictionary.

        \subsection{Effects of Weighting Parameter $\lambda$}
            \label{sec:experiment_lambda}

            Parameter $\lambda$ balances the Global Position Cost and Local Adjacency Cost in the objective function. To examine the effect of this balancing, we set $\lambda$ to $0.0, 0.1, 0.2, .., 1.0$ and reconstructed images.

            Fig.~\ref{fig:experiment_lambda} presents the results. If Local Adjacency Costs are excluded ($\lambda = 0.0$), the reconstructed images have faint shapes, but they are heavily blurred and are difficult to identify. However, if Global Position Cost is excluded ($\lambda = 1.0$), then reconstructed images tend to have corrupted shapes although they have clear contours. Those results indicate that the costs of both types are working effectively and that they must be balanced properly.

\begin{figure}
                \begin{center}
                    \includegraphics[bb=0 0 432 252,width=1.0\linewidth]{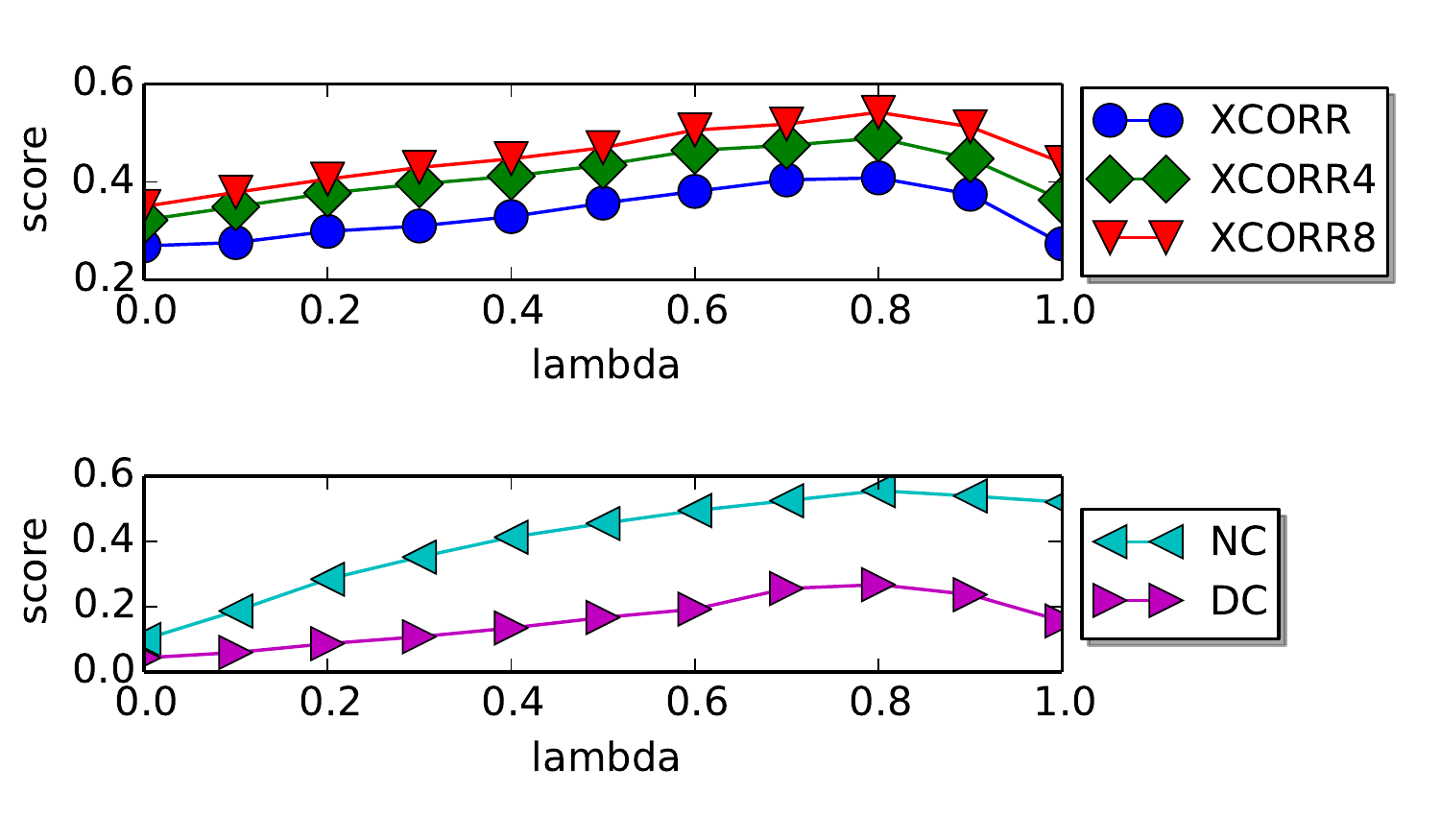}
                \end{center}
                \caption{Quantitative evaluation of the effects of parameter $\lambda$. The scores are averaged over $101$ images. Metrics are defined in Section \ref{sec:method_evaluation}. For all metrics, higher values are better.}
                \label{fig:lambda_quantitative}
            \end{figure}

           Fig.~\ref{fig:lambda_quantitative} presents the relation between $\lambda$ and reconstruction accuracy. The best results are produced at $\lambda=0.8$ for all evaluation metrics. It is interesting that NC also takes the highest value at $\lambda=0.8$, not at $\lambda=1.0$.

        \subsection{Effect of Dictionary Size}
        \label{sec:experiment_quantization}

            \begin{figure*}
                \begin{center}
                    \raisebox{5mm}{\makebox[28mm][l]{\small{(a) $k=128$}}}
                    \includegraphics[width=13.4mm,bb=0 0 128 128]{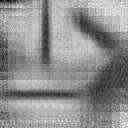}
                    \includegraphics[width=13.4mm,bb=0 0 128 128]{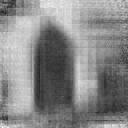}
                    \includegraphics[width=13.4mm,bb=0 0 128 128]{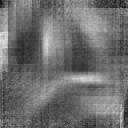}
                    \includegraphics[width=13.4mm,bb=0 0 128 128]{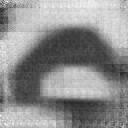}
                    \includegraphics[width=13.4mm,bb=0 0 128 128]{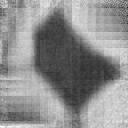}
                    \includegraphics[width=13.4mm,bb=0 0 128 128]{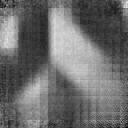}
                    \includegraphics[width=13.4mm,bb=0 0 128 128]{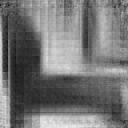}
                    \includegraphics[width=13.4mm,bb=0 0 128 128]{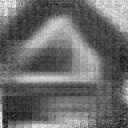}
                    \includegraphics[width=13.4mm,bb=0 0 128 128]{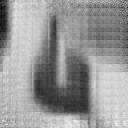}
                    \includegraphics[width=13.4mm,bb=0 0 128 128]{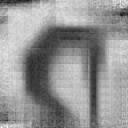} \\
                    \raisebox{5mm}{\makebox[28mm][l]{\small{(b) $k=512$}}}
                    \includegraphics[width=13.4mm,bb=0 0 128 128]{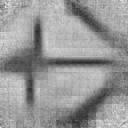}
                    \includegraphics[width=13.4mm,bb=0 0 128 128]{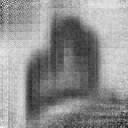}
                    \includegraphics[width=13.4mm,bb=0 0 128 128]{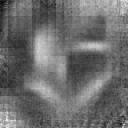}
                    \includegraphics[width=13.4mm,bb=0 0 128 128]{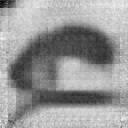}
                    \includegraphics[width=13.4mm,bb=0 0 128 128]{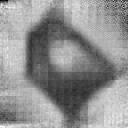}
                    \includegraphics[width=13.4mm,bb=0 0 128 128]{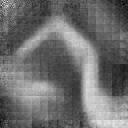}
                    \includegraphics[width=13.4mm,bb=0 0 128 128]{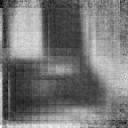}
                    \includegraphics[width=13.4mm,bb=0 0 128 128]{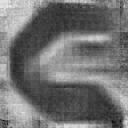}
                    \includegraphics[width=13.4mm,bb=0 0 128 128]{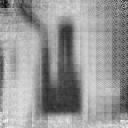}
                    \includegraphics[width=13.4mm,bb=0 0 128 128]{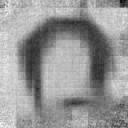} \\
                    \raisebox{5mm}{\makebox[28mm][l]{\small{(c) $k=2048$}}}
                    \includegraphics[width=13.4mm,bb=0 0 128 128]{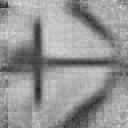}
                    \includegraphics[width=13.4mm,bb=0 0 128 128]{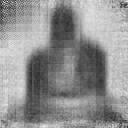}
                    \includegraphics[width=13.4mm,bb=0 0 128 128]{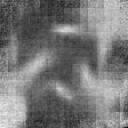}
                    \includegraphics[width=13.4mm,bb=0 0 128 128]{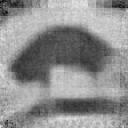}
                    \includegraphics[width=13.4mm,bb=0 0 128 128]{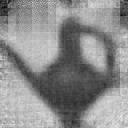}
                    \includegraphics[width=13.4mm,bb=0 0 128 128]{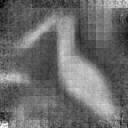}
                    \includegraphics[width=13.4mm,bb=0 0 128 128]{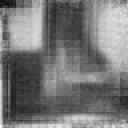}
                    \includegraphics[width=13.4mm,bb=0 0 128 128]{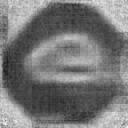}
                    \includegraphics[width=13.4mm,bb=0 0 128 128]{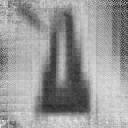}
                    \includegraphics[width=13.4mm,bb=0 0 128 128]{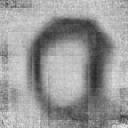} \\
                    \raisebox{5mm}{\makebox[28mm][l]{\small{(d) $k=8192$}}}
                    \includegraphics[width=13.4mm,bb=0 0 128 128]{./data/proposed/SIFT_k_8192_lambda_08/007.jpg}
                    \includegraphics[width=13.4mm,bb=0 0 128 128]{./data/proposed/SIFT_k_8192_lambda_08/016.jpg}
                    \includegraphics[width=13.4mm,bb=0 0 128 128]{./data/proposed/SIFT_k_8192_lambda_08/096.jpg}
                    \includegraphics[width=13.4mm,bb=0 0 128 128]{./data/proposed/SIFT_k_8192_lambda_08/038.jpg}
                    \includegraphics[width=13.4mm,bb=0 0 128 128]{./data/proposed/SIFT_k_8192_lambda_08/040.jpg}
                    \includegraphics[width=13.4mm,bb=0 0 128 128]{./data/proposed/SIFT_k_8192_lambda_08/052.jpg}
                    \includegraphics[width=13.4mm,bb=0 0 128 128]{./data/proposed/SIFT_k_8192_lambda_08/023.jpg}
                    \includegraphics[width=13.4mm,bb=0 0 128 128]{./data/proposed/SIFT_k_8192_lambda_08/010.jpg}
                    \includegraphics[width=13.4mm,bb=0 0 128 128]{./data/proposed/SIFT_k_8192_lambda_08/065.jpg}
                    \includegraphics[width=13.4mm,bb=0 0 128 128]{./data/proposed/SIFT_k_8192_lambda_08/002.jpg} \\
                \end{center}
                \caption{Examples of reconstructed images. The number of visual words $k$ varies.}
                \label{fig:parameter_k}
            \end{figure*}

            As demonstrated in Section~\ref{sec:experiment_aligned}, the fact that local descriptors are quantized to visual words has little impact in converting local descriptors into image patches. However, it can affect the estimation of their spatial arrangement. To examine the related effects, we changed the size of visual word dictionary $k$ and reconstructed images.

            Fig.~\ref{fig:parameter_k} presents the results. Images are fine if $k \geq2048$. However, shapes of the objects tend to be corrupted if $k \leq 512$. From the images of a chair in the seventh column, it is apparent that the chair becomes invisible at $k = 512$, which indicates the possibility that the feature cannot capture the chair. Analysis of this type is probably useful for tuning $k$.
            
\begin{figure}
                \begin{center}
                    \includegraphics[bb=0 0 432 252,width=1.0\linewidth]{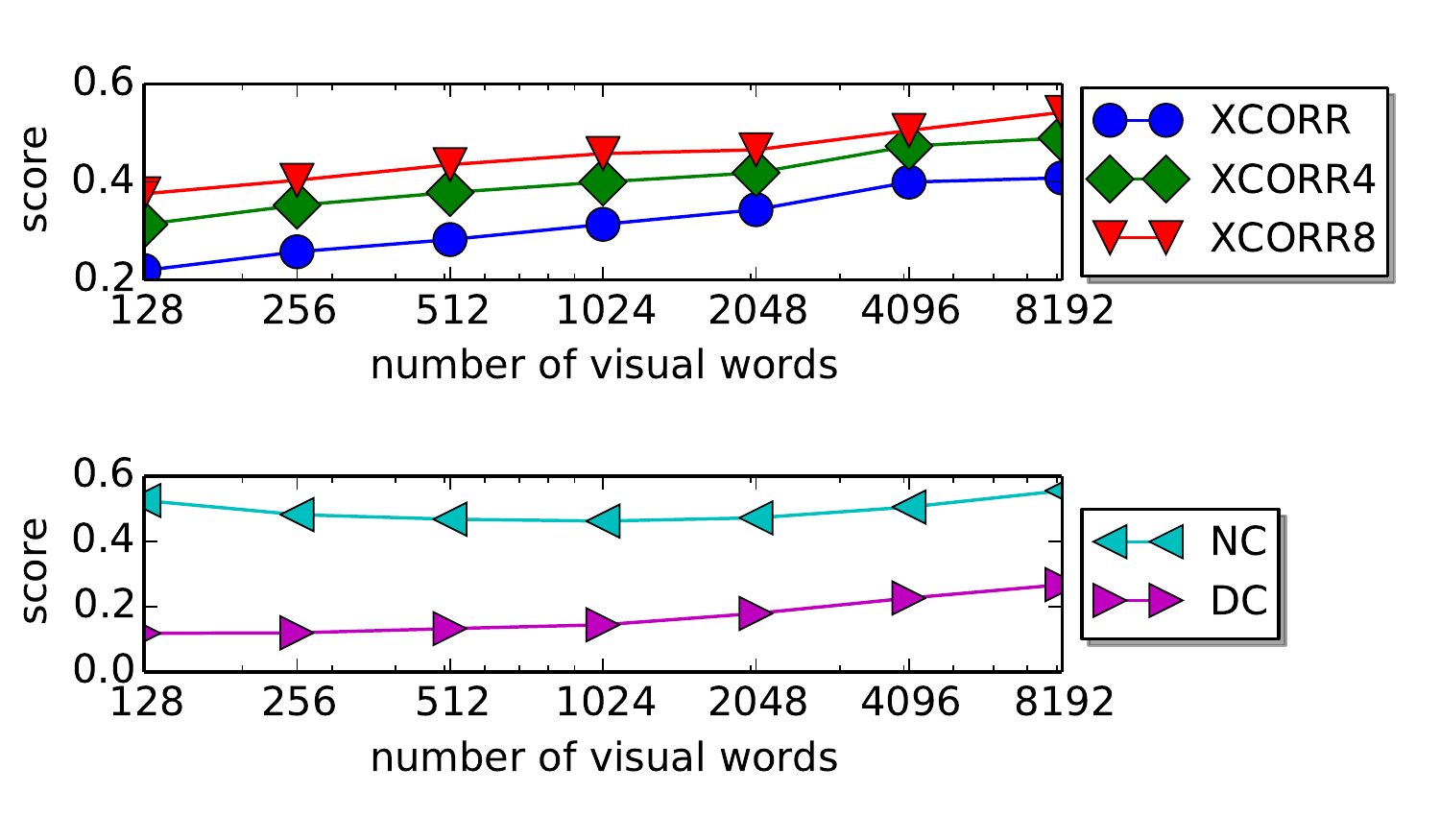}
                \end{center}
                \caption{Relation between the number of visual words and reconstruction score. The scores are averaged over $101$ images. Metrics are defined in Section \ref{sec:method_evaluation}. For all metrics, larger values are better.}
                \label{fig:k_quantitative}
            \end{figure}

            Fig.~\ref{fig:k_quantitative} shows the relation between $k$ and evaluation metrics. As expected, larger $k$ produces larger XCORR, XCORR4, and XCORR8. Interpreting NC and DC is rather difficult because multiple descriptors are quantized to one visual word and the chance rate of correct assignment becomes higher when the dictionary becomes smaller.

        \subsection{Comparison of Optimization Methods}
            \label{sec:experiment_optimization}

            In this section, we compare our optimization method (HC+GA) with random assignment (RAND), Hill Climbing (HC), and simulated annealing (SA). The way to modify a solution in HC and SA is the same as that described in Section~\ref{sec:method_optimization}. We reconstructed $101$ images using each optimization method and computed the evaluation metrics described in Section~\ref{sec:method_evaluation}.

            Table~\ref{table:experiment_optimization} presents evaluation values, optimized values of the objective function, and the running time to finish optimization. Our method outperforms other methods at all metrics. The most significant difference among them is the ratio of visual words assigned to the correct position (DC), which is seven times as large as HC, and four times as large as SA. However, our method requires much more time than other methods do.

            \begin{table}
                \caption{Comparison of optimization methods. RAND is a randomly generated solution. HC denotes Hill Climbing. SA represents simulated annealing. GA+HC is a combination of genetic algorithm and Hill Climbing.}
                \begin{center}
                    \begin{tabular}{|l|c|c|c|c|}
                        \hline
                        method & RAND & HC & SA & GA+HC \\
                        \hline\hline
                        XCORR             & 0.037  & 0.165 & 0.229 & 0.426 \\
                        XCORR4            & 0.111  & 0.257 & 0.342 & 0.495 \\
                        XCORR8            & 0.176  & 0.332 & 0.431 & 0.532 \\
                        \hline
                        NC                & 0.016  & 0.327 & 0.419 & 0.541 \\
                        DC                & 0.014  & 0.036 & 0.064 & 0.259 \\
                        \hline
                        value of Eq.~\ref{eq:cost} & 956.9  & 740.9 & 704.5 & 683.3 \\
                        \hline
                        time [s]        & 0.0004 & 0.102 & 4.239 & 51.15 \\
                        \hline
                    \end{tabular}
                \end{center}
                \label{table:experiment_optimization}
            \end{table}

        \subsection{Image Reconstruction from Various Descriptors}
            \label{sec:experiment_various_descriptors}

            \begin{figure*}
              \begin{center}
                \raisebox{5mm}{\makebox[28mm][l]{\small{(a) RGBSIFT}}}
                \includegraphics[width=13.4mm,bb=0 0 128 128]{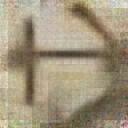}
                \includegraphics[width=13.4mm,bb=0 0 128 128]{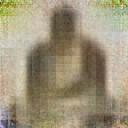}
                \includegraphics[width=13.4mm,bb=0 0 128 128]{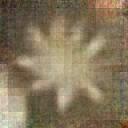}
                \includegraphics[width=13.4mm,bb=0 0 128 128]{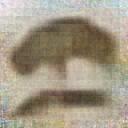}
                \includegraphics[width=13.4mm,bb=0 0 128 128]{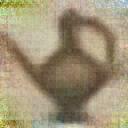}
                \includegraphics[width=13.4mm,bb=0 0 128 128]{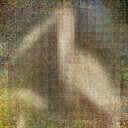}
                \includegraphics[width=13.4mm,bb=0 0 128 128]{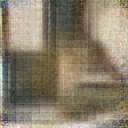}
                \includegraphics[width=13.4mm,bb=0 0 128 128]{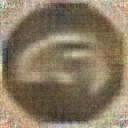}
                \includegraphics[width=13.4mm,bb=0 0 128 128]{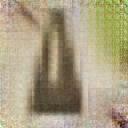}
                \includegraphics[width=13.4mm,bb=0 0 128 128]{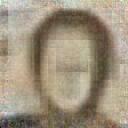} \\

                \raisebox{5mm}{\makebox[28mm][l]{\small{(b) OpponentSIFT}}}
                \includegraphics[width=13.4mm,bb=0 0 128 128]{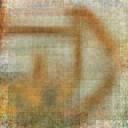}
                \includegraphics[width=13.4mm,bb=0 0 128 128]{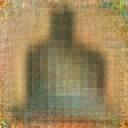}
                \includegraphics[width=13.4mm,bb=0 0 128 128]{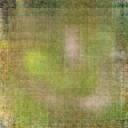}
                \includegraphics[width=13.4mm,bb=0 0 128 128]{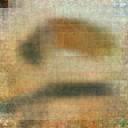}
                \includegraphics[width=13.4mm,bb=0 0 128 128]{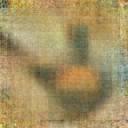}
                \includegraphics[width=13.4mm,bb=0 0 128 128]{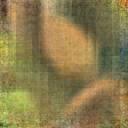}
                \includegraphics[width=13.4mm,bb=0 0 128 128]{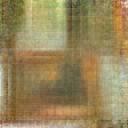}
                \includegraphics[width=13.4mm,bb=0 0 128 128]{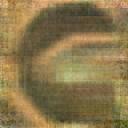}
                \includegraphics[width=13.4mm,bb=0 0 128 128]{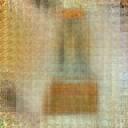}
                \includegraphics[width=13.4mm,bb=0 0 128 128]{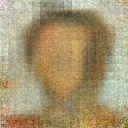} \\
                \raisebox{5mm}{\makebox[28mm][l]{\small{(c) HOG}}}
                \includegraphics[width=13.4mm,bb=0 0 128 128]{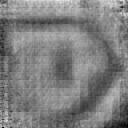}
                \includegraphics[width=13.4mm,bb=0 0 128 128]{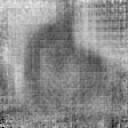}
                \includegraphics[width=13.4mm,bb=0 0 128 128]{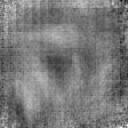}
                \includegraphics[width=13.4mm,bb=0 0 128 128]{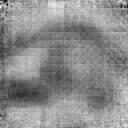}
                \includegraphics[width=13.4mm,bb=0 0 128 128]{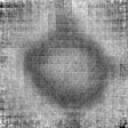}
                \includegraphics[width=13.4mm,bb=0 0 128 128]{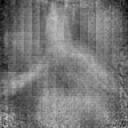}
                \includegraphics[width=13.4mm,bb=0 0 128 128]{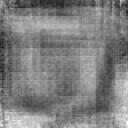}
                \includegraphics[width=13.4mm,bb=0 0 128 128]{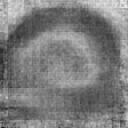}
                \includegraphics[width=13.4mm,bb=0 0 128 128]{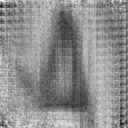}
                \includegraphics[width=13.4mm,bb=0 0 128 128]{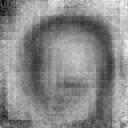} \\
                \raisebox{5mm}{\makebox[28mm][l]{\small{(d) LBP}}}
                \includegraphics[width=13.4mm,bb=0 0 128 128]{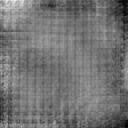}
                \includegraphics[width=13.4mm,bb=0 0 128 128]{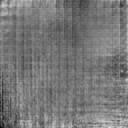}
                \includegraphics[width=13.4mm,bb=0 0 128 128]{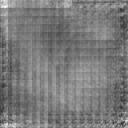}
                \includegraphics[width=13.4mm,bb=0 0 128 128]{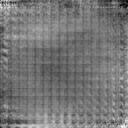}
                \includegraphics[width=13.4mm,bb=0 0 128 128]{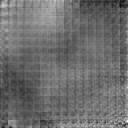}
                \includegraphics[width=13.4mm,bb=0 0 128 128]{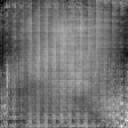}
                \includegraphics[width=13.4mm,bb=0 0 128 128]{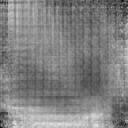}
                \includegraphics[width=13.4mm,bb=0 0 128 128]{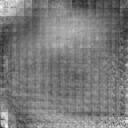}
                \includegraphics[width=13.4mm,bb=0 0 128 128]{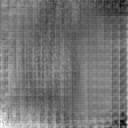}
                \includegraphics[width=13.4mm,bb=0 0 128 128]{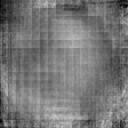} \\
              \end{center}
              \caption{Reconstructed images. Descriptors of various kinds are used. }
              \label{fig:various_descriptors}
            \end{figure*}

            Although SIFT descriptors were used in preceding experiments, our method is applicable to descriptors of any kind. Reconstructed images might vary along with the kind of descriptor, and exhibit their characteristics. Here we reconstructed images using RGBSIFT~\cite{van2008comparison}, OpponentSIFT~\cite{van2008comparison}, HOG~\cite{dalal2005histograms}, and LBP~\cite{ojala1996comparative} as local descriptors. Results are shown in Fig.~\ref{fig:various_descriptors}.

            Images of RGBSIFT and OpponentSIFT are largely well reconstructed. They have colors because their descriptors include color information. Images of RGBSIFT are more accurate than images of OpponentSIFT. However, RGBSIFT has less capacity for color representation than OpponentSIFT has. From this observation, we can understand that there exists a tradeoff between color representation and edge representation. Although shapes and edges are reconstructed correctly, the color variation of these images is awfully limited. Most of their colors are not consistent with those of the original images in Fig.~\ref{fig:reconstruced_from_bovw}, which indicates that they cannot capture color information adequately, probably because the dictionary of visual words is too small or biased.

            Reconstructed images of HOG are much worse than those of SIFT, although they exhibit partial edges. 

            Most images using LBP are extremely obscure and show no clear edges, which indicates that LBP cannot capture edge information well.

    \section{Applications of Proposed Method}
        \label{chap:application}

        Our method is useful not only for reconstructing images, but also for generating novel images from BoVWs manipulated in the feature space. In this section, we present three image-generation applications. The first is image morphing by interpolating two features. The second is visualizing learned object classifiers. The third is image generation from natural sentences.
        
        \subsection{Morphing of Images}
            \label{sec:application_morphing}

            Morphing is a special effect that changes one image into another through a seamless transition. Although making intermediate images directly is not straightforward, it is rather simple to generate images from their intermediate features. Therefore, in this section, we apply our method to image morphing. 

            The feature space of BoVW is preferred for recognition and retrieval more than raw pixels. Therefore, image morphing via image features can be more semantic than conventional approaches. Image morphing via interpolation in the feature space is used to examine the manifold property of features in representation learning~\cite{bengio2012better, kingma2013auto}.

            \subsubsection{Method}

                Let $W_s$ and $W_t$ represent the sets of visual words extracted from image $s$ and image $t$. We obtain intermediate BoVWs between $W_s$ and $W_t$ and morph from $s$ to $t$ by the following procedure. 
                \begin{enumerate}
                    \item Let $ i \leftarrow 1, W_1 \leftarrow W_s$. 
                    \item Let $i \leftarrow i+1, W_i \leftarrow W_{i-1}$. 
                    \item Select one visual word randomly from $W_t \cap \overline{W_i}$ and append it to $W_i$. 
                    \item Select one visual word randomly from $W_i \cap \overline{W_t}$ and remove it from $W_i$. 
                    \item Output $W_1, ..., W_i$ if $W_i = W_t$. Go to 2) otherwise. 
                    \item Generate images from BoVW $W_1, ..., W_i$.
                \end{enumerate}

            \subsubsection{Experiment}

                \begin{figure*}[t]
                  \begin{center}
                    \raisebox{5mm}{\makebox[8mm][l]{\small{(a)}}}
                    \includegraphics[width=13.4mm,bb=0 0 128 128]{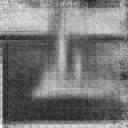}
                    \includegraphics[width=13.4mm,bb=0 0 128 128]{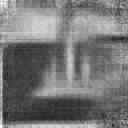}
                    \includegraphics[width=13.4mm,bb=0 0 128 128]{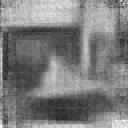}
                    \includegraphics[width=13.4mm,bb=0 0 128 128]{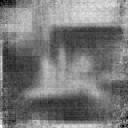}
                    \includegraphics[width=13.4mm,bb=0 0 128 128]{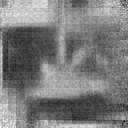}
                    \includegraphics[width=13.4mm,bb=0 0 128 128]{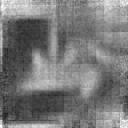}
                    \includegraphics[width=13.4mm,bb=0 0 128 128]{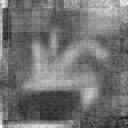}
                    \includegraphics[width=13.4mm,bb=0 0 128 128]{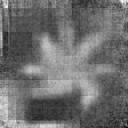}
                    \includegraphics[width=13.4mm,bb=0 0 128 128]{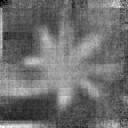}
                    \includegraphics[width=13.4mm,bb=0 0 128 128]{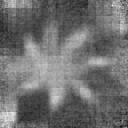}
                    \includegraphics[width=13.4mm,bb=0 0 128 128]{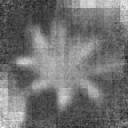} \\
                    \raisebox{5mm}{\makebox[8mm][l]{\small{(b)}}}
                    \includegraphics[width=13.4mm,bb=0 0 128 128]{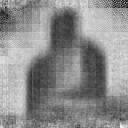}
                    \includegraphics[width=13.4mm,bb=0 0 128 128]{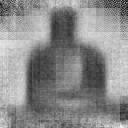}
                    \includegraphics[width=13.4mm,bb=0 0 128 128]{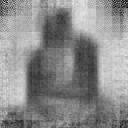}
                    \includegraphics[width=13.4mm,bb=0 0 128 128]{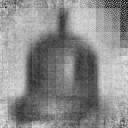}
                    \includegraphics[width=13.4mm,bb=0 0 128 128]{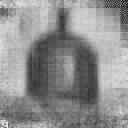}
                    \includegraphics[width=13.4mm,bb=0 0 128 128]{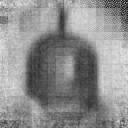}
                    \includegraphics[width=13.4mm,bb=0 0 128 128]{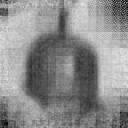}
                    \includegraphics[width=13.4mm,bb=0 0 128 128]{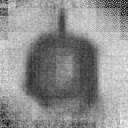}
                    \includegraphics[width=13.4mm,bb=0 0 128 128]{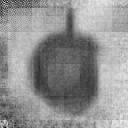}
                    \includegraphics[width=13.4mm,bb=0 0 128 128]{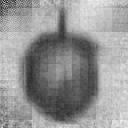}
                    \includegraphics[width=13.4mm,bb=0 0 128 128]{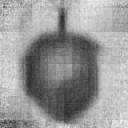} \\
                    \raisebox{5mm}{\makebox[8mm][l]{\small{(c)}}}
                    \includegraphics[width=13.4mm,bb=0 0 128 128]{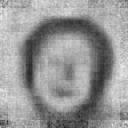}
                    \includegraphics[width=13.4mm,bb=0 0 128 128]{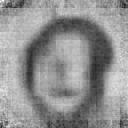}
                    \includegraphics[width=13.4mm,bb=0 0 128 128]{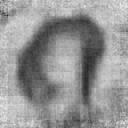}
                    \includegraphics[width=13.4mm,bb=0 0 128 128]{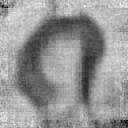}
                    \includegraphics[width=13.4mm,bb=0 0 128 128]{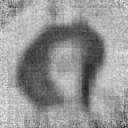}
                    \includegraphics[width=13.4mm,bb=0 0 128 128]{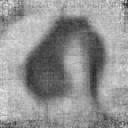}
                    \includegraphics[width=13.4mm,bb=0 0 128 128]{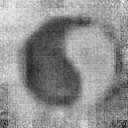}
                    \includegraphics[width=13.4mm,bb=0 0 128 128]{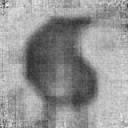}
                    \includegraphics[width=13.4mm,bb=0 0 128 128]{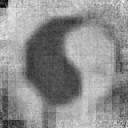}
                    \includegraphics[width=13.4mm,bb=0 0 128 128]{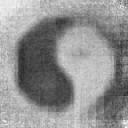}
                    \includegraphics[width=13.4mm,bb=0 0 128 128]{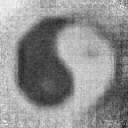} \\
                    \raisebox{5mm}{\makebox[8mm][l]{\small{(d)}}}
                    \includegraphics[width=13.4mm,bb=0 0 128 128]{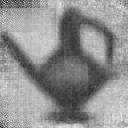}
                    \includegraphics[width=13.4mm,bb=0 0 128 128]{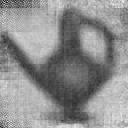}
                    \includegraphics[width=13.4mm,bb=0 0 128 128]{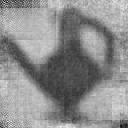}
                    \includegraphics[width=13.4mm,bb=0 0 128 128]{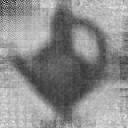}
                    \includegraphics[width=13.4mm,bb=0 0 128 128]{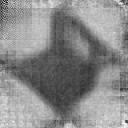}
                    \includegraphics[width=13.4mm,bb=0 0 128 128]{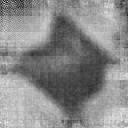}
                    \includegraphics[width=13.4mm,bb=0 0 128 128]{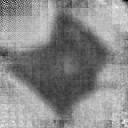}
                    \includegraphics[width=13.4mm,bb=0 0 128 128]{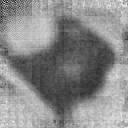}
                    \includegraphics[width=13.4mm,bb=0 0 128 128]{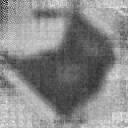}
                    \includegraphics[width=13.4mm,bb=0 0 128 128]{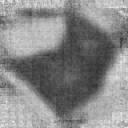}
                    \includegraphics[width=13.4mm,bb=0 0 128 128]{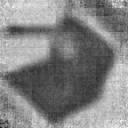} \\
                  \end{center}
                  \caption{Morphing by image generation from BoVW. BoVWs extracted from two images are interpolated and converted to images. (a) A ketch to a sunflower. (b) Buddha to a strawberry. (c) Face to a yin--yang. (d) Ewer to a laptop.}
                  \label{fig:morphing}
                \end{figure*}

                Fig.~\ref{fig:morphing} shows examples of images generated by the preceding procedure. Most intermediate images are omitted because of limited space. The results showed, apparently, that one object gradually transforms another object. Although they are unsuitable for practical use by now because of low image quality, image editing via image features is an interesting avenue.

        \subsection{Visualization of Object Classifiers}
            \label{sec:application_classifier}

             In this section, we visualize object classifiers by generating images from features that activate the classifiers the most. This enables us to understand intuitively what the classifiers captured. This also reveals differences between human vision and computer vision, as suggested for Deep Neural Networks~\cite{nguyen2014deep}. We used linear SVMs as object classifiers because they are commonly used as the first choice to classify BoVWs.

            \subsubsection{Method}

                A linear classifier has weight vector $\bm{w}$ and bias parameter $b$. Letting $\bm{x}$ be a feature vector, then the linear classifier computes the following score $y$.
                \begin{align}
                    y = \bm{w}^T \bm{x} + b \; .
                \end{align}
                If $y$ is positive, then $\bm{x}$ is judged to belong to positive class.

                Let us consider feature $\bm{x}^*$ that maximizes score $y$. If $\bm{x}$ is multiplied, then $y$ is also multiplied. Therefore $y$ can be infinity if no limitation exists on $\bm{x}$. Therefore, we restrict $\bm{x}$ to $ \|\bm{x}\| = 1 $. Then we get $\bm{x}^* = \bm{w} / \| \bm{w} \|$ by Lagrange's method of undetermined multipliers.

                $\bm{x}^*$ is difficult to interpret as a "bag of visual words" because each dimension of $\bm{x}^*$ can be a negative or non-integer. Therefore we modify $\bm{x}^*$ that each dimension of it is non-negative integer and L1 norm of $\bm{x}^*$ is the number of visual words in an image $n$. Firstly we set a negative value of $\bm{x}^*$ to $0$. Secondly, we find out $\alpha$ which satisfies $ \|\mbox{round}(\alpha \bm{x}^*) \|_1 = n$ and rescale $x^{**} = \mbox{round}(\alpha \bm{x}^*)$. This procedure enables us to interpret $x^{**}$ as a set of $n$ visual words.

            \subsubsection{Experiment}

                \begin{figure*}
                    \centering
                    \subfloat[bonsai]{\makebox[18mm][c]{\includegraphics[width=13.4mm,bb=0 0 128 128]{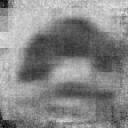}}}
                    \subfloat[brontosaurus]{\makebox[18mm][c]{\includegraphics[width=13.4mm,bb=0 0 128 128]{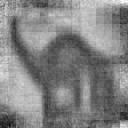}}}
                    \subfloat[camera]{\makebox[18mm][c]{\includegraphics[width=13.4mm,bb=0 0 128 128]{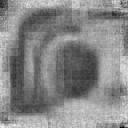}}}
                    \subfloat[dragonfly]{\makebox[18mm][c]{\includegraphics[width=13.4mm,bb=0 0 128 128]{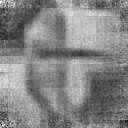}}}
                    \subfloat[faces]{\makebox[18mm][c]{\includegraphics[width=13.4mm,bb=0 0 128 128]{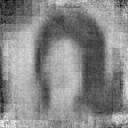}}}
                    \subfloat[ferry]{\makebox[18mm][c]{\includegraphics[width=13.4mm,bb=0 0 128 128]{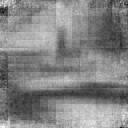}}}
                    \subfloat[gramophone]{\makebox[18mm][c]{\includegraphics[width=13.4mm,bb=0 0 128 128]{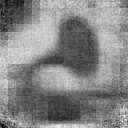}}}
                    \subfloat[grand piano]{\makebox[18mm][c]{\includegraphics[width=13.4mm,bb=0 0 128 128]{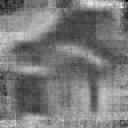}}}
                    \subfloat[headphone]{\makebox[18mm][c]{\includegraphics[width=13.4mm,bb=0 0 128 128]{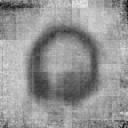}}} \\
                    \subfloat[joshua tree]{\makebox[18mm][c]{\includegraphics[width=13.4mm,bb=0 0 128 128]{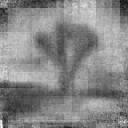}}}
                    \subfloat[laptop]{\makebox[18mm][c]{\includegraphics[width=13.4mm,bb=0 0 128 128]{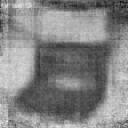}}}
                    \subfloat[pizza]{\makebox[18mm][c]{\includegraphics[width=13.4mm,bb=0 0 128 128]{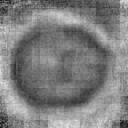}}}
                    \subfloat[pyramid]{\makebox[18mm][c]{\includegraphics[width=13.4mm,bb=0 0 128 128]{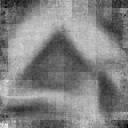}}}
                    \subfloat[revolver]{\makebox[18mm][c]{\includegraphics[width=13.4mm,bb=0 0 128 128]{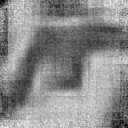}}}
                    \subfloat[schooner]{\makebox[18mm][c]{\includegraphics[width=13.4mm,bb=0 0 128 128]{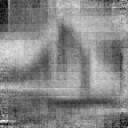}}}
                    \subfloat[tick]{\makebox[18mm][c]{\includegraphics[width=13.4mm,bb=0 0 128 128]{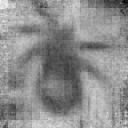}}}
                    \subfloat[umbrella]{\makebox[18mm][c]{\includegraphics[width=13.4mm,bb=0 0 128 128]{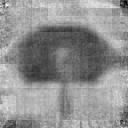}}}
                    \subfloat[wheelchair]{\makebox[18mm][c]{\includegraphics[width=13.4mm,bb=0 0 128 128]{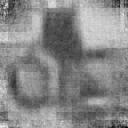}}}
                    \caption{Visualized object classifiers. \label{fig:classifier}}
                \end{figure*}

                We trained $101$ SVM classifiers with linear kernels using the Caltech $101$ Dataset and the ILSVRC $2012$ Dataset. All images in each class of the Caltech $101$ Dataset were treated as positive examples. Ten thousand images randomly extracted from ILSVRC $2012$ Dataset were treated as negative examples. We generated images from each classifier five times and selected the clearest image manually because the optimization of layouts sometimes became stuck at poor local optima.

\begin{figure*}
            \centering
            \parbox[c][16.0mm][t]{9.6mm}{\includegraphics[width=9.6mm,bb=0 0 128 128]{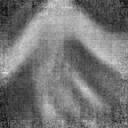} \newline \scriptsize{accor\-dion}}
            \parbox[c][16.0mm][t]{9.6mm}{\includegraphics[width=9.6mm,bb=0 0 128 128]{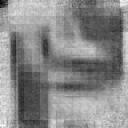} \newline \scriptsize{air\-planes}}
            \parbox[c][16.0mm][t]{9.6mm}{\includegraphics[width=9.6mm,bb=0 0 128 128]{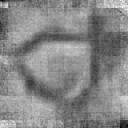} \newline \scriptsize{anchor}}
            \parbox[c][16.0mm][t]{9.6mm}{\includegraphics[width=9.6mm,bb=0 0 128 128]{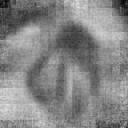} \newline \scriptsize{ant}}
            \parbox[c][16.0mm][t]{9.6mm}{\includegraphics[width=9.6mm,bb=0 0 128 128]{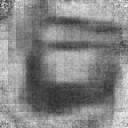} \newline \scriptsize{barrel}}
            \parbox[c][16.0mm][t]{9.6mm}{\includegraphics[width=9.6mm,bb=0 0 128 128]{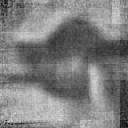} \newline \scriptsize{bass}}
            \parbox[c][16.0mm][t]{9.6mm}{\includegraphics[width=9.6mm,bb=0 0 128 128]{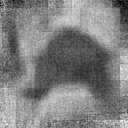} \newline \scriptsize{beaver}}
            \parbox[c][16.0mm][t]{9.6mm}{\includegraphics[width=9.6mm,bb=0 0 128 128]{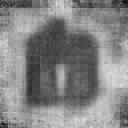} \newline \scriptsize{binocu\-lar}}
            \parbox[c][16.0mm][t]{9.6mm}{\includegraphics[width=9.6mm,bb=0 0 128 128]{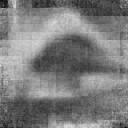} \newline \scriptsize{bonsai}}
            \parbox[c][16.0mm][t]{9.6mm}{\includegraphics[width=9.6mm,bb=0 0 128 128]{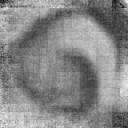} \newline \scriptsize{brain}}
            \parbox[c][16.0mm][t]{9.6mm}{\includegraphics[width=9.6mm,bb=0 0 128 128]{./data/classifier/brontosaurus_08_3.jpg} \newline \scriptsize{bronto\-saurus}}
            \parbox[c][16.0mm][t]{9.6mm}{\includegraphics[width=9.6mm,bb=0 0 128 128]{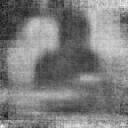} \newline \scriptsize{buddha}}
            \parbox[c][16.0mm][t]{9.6mm}{\includegraphics[width=9.6mm,bb=0 0 128 128]{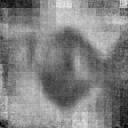} \newline \scriptsize{butter- \\ fly}}
            \parbox[c][16.0mm][t]{9.6mm}{\includegraphics[width=9.6mm,bb=0 0 128 128]{./data/classifier/camera_08_4.jpg} \newline \scriptsize{camera}}
            \parbox[c][16.0mm][t]{9.6mm}{\includegraphics[width=9.6mm,bb=0 0 128 128]{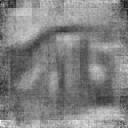} \newline \scriptsize{cannon}}
            \parbox[c][16.0mm][t]{9.6mm}{\includegraphics[width=9.6mm,bb=0 0 128 128]{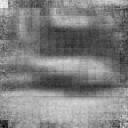} \newline \scriptsize{car side}}
            \parbox[c][16.0mm][t]{9.6mm}{\includegraphics[width=9.6mm,bb=0 0 128 128]{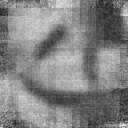} \newline \scriptsize{ceiling fan}}
            \parbox[c][16.0mm][t]{9.6mm}{\includegraphics[width=9.6mm,bb=0 0 128 128]{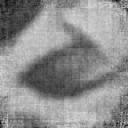} \newline \scriptsize{cell\-phone}}
            \parbox[c][16.0mm][t]{9.6mm}{\includegraphics[width=9.6mm,bb=0 0 128 128]{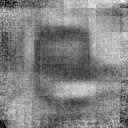} \newline \scriptsize{chair}}
            \parbox[c][16.0mm][t]{9.6mm}{\includegraphics[width=9.6mm,bb=0 0 128 128]{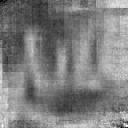} \newline \scriptsize{chan\-de\-lier}}
            \parbox[c][16.0mm][t]{9.6mm}{\includegraphics[width=9.6mm,bb=0 0 128 128]{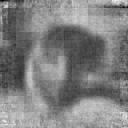} \newline \scriptsize{cougar body}}
            \parbox[c][16.0mm][t]{9.6mm}{\includegraphics[width=9.6mm,bb=0 0 128 128]{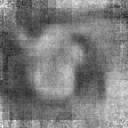} \newline \scriptsize{cougar face}}
            \parbox[c][16.0mm][t]{9.6mm}{\includegraphics[width=9.6mm,bb=0 0 128 128]{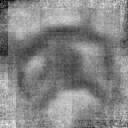} \newline \scriptsize{crab}}
            \parbox[c][16.0mm][t]{9.6mm}{\includegraphics[width=9.6mm,bb=0 0 128 128]{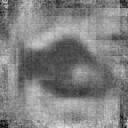} \newline \scriptsize{crayfish}}
            \parbox[c][16.0mm][t]{9.6mm}{\includegraphics[width=9.6mm,bb=0 0 128 128]{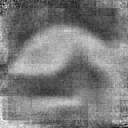} \newline \scriptsize{croco\-dile}}
            \parbox[c][16.0mm][t]{9.6mm}{\includegraphics[width=9.6mm,bb=0 0 128 128]{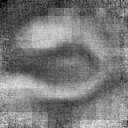} \newline \scriptsize{crocodile head}}
            \parbox[c][16.0mm][t]{9.6mm}{\includegraphics[width=9.6mm,bb=0 0 128 128]{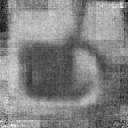} \newline \scriptsize{cup}}
            \parbox[c][16.0mm][t]{9.6mm}{\includegraphics[width=9.6mm,bb=0 0 128 128]{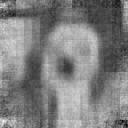} \newline \scriptsize{dal\-ma\-tian}}
            \parbox[c][16.0mm][t]{9.6mm}{\includegraphics[width=9.6mm,bb=0 0 128 128]{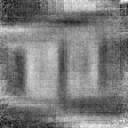} \newline \scriptsize{dollar bill}}
            \parbox[c][16.0mm][t]{9.6mm}{\includegraphics[width=9.6mm,bb=0 0 128 128]{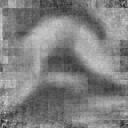} \newline \scriptsize{dolphin}}
            \parbox[c][16.0mm][t]{9.6mm}{\includegraphics[width=9.6mm,bb=0 0 128 128]{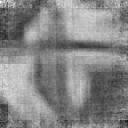} \newline \scriptsize{dragon\-fly}}
            \parbox[c][16.0mm][t]{9.6mm}{\includegraphics[width=9.6mm,bb=0 0 128 128]{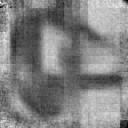} \newline \scriptsize{electric guitar}}
            \parbox[c][16.0mm][t]{9.6mm}{\includegraphics[width=9.6mm,bb=0 0 128 128]{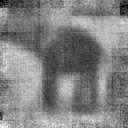} \newline \scriptsize{ele\-phant}}
            \parbox[c][16.0mm][t]{9.6mm}{\includegraphics[width=9.6mm,bb=0 0 128 128]{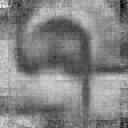} \newline \scriptsize{emu}}
            \parbox[c][16.0mm][t]{9.6mm}{\includegraphics[width=9.6mm,bb=0 0 128 128]{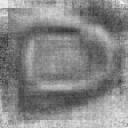} \newline \scriptsize{eu\-pho\-ni\-um}}
            \parbox[c][16.0mm][t]{9.6mm}{\includegraphics[width=9.6mm,bb=0 0 128 128]{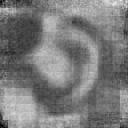} \newline \scriptsize{ewer}}
            \parbox[c][16.0mm][t]{9.6mm}{\includegraphics[width=9.6mm,bb=0 0 128 128]{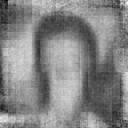} \newline \scriptsize{faces}}
            \parbox[c][16.0mm][t]{9.6mm}{\includegraphics[width=9.6mm,bb=0 0 128 128]{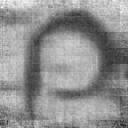} \newline \scriptsize{faces easy}}
            \parbox[c][16.0mm][t]{9.6mm}{\includegraphics[width=9.6mm,bb=0 0 128 128]{./data/classifier/ferry_08_3.jpg} \newline \scriptsize{ferry}}
            \parbox[c][16.0mm][t]{9.6mm}{\includegraphics[width=9.6mm,bb=0 0 128 128]{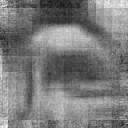} \newline \scriptsize{flamingo}}
            \parbox[c][16.0mm][t]{9.6mm}{\includegraphics[width=9.6mm,bb=0 0 128 128]{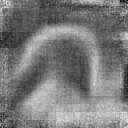} \newline \scriptsize{flamingo head}}
            \parbox[c][16.0mm][t]{9.6mm}{\includegraphics[width=9.6mm,bb=0 0 128 128]{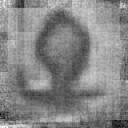} \newline \scriptsize{garfield}}
            \parbox[c][16.0mm][t]{9.6mm}{\includegraphics[width=9.6mm,bb=0 0 128 128]{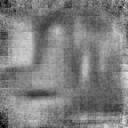} \newline \scriptsize{gerenuk}}
            \parbox[c][16.0mm][t]{9.6mm}{\includegraphics[width=9.6mm,bb=0 0 128 128]{./data/classifier/gramophone_08_4.jpg} \newline \scriptsize{gram\-o\-phone}}
            \parbox[c][16.0mm][t]{9.6mm}{\includegraphics[width=9.6mm,bb=0 0 128 128]{./data/classifier/grand_piano_08_2.jpg} \newline \scriptsize{grand piano}}
            \parbox[c][16.0mm][t]{9.6mm}{\includegraphics[width=9.6mm,bb=0 0 128 128]{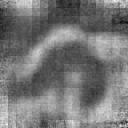} \newline \scriptsize{hawks\-bill}}
            \parbox[c][16.0mm][t]{9.6mm}{\includegraphics[width=9.6mm,bb=0 0 128 128]{./data/classifier/headphone_08_5.jpg} \newline \scriptsize{head\-phone}}
            \parbox[c][16.0mm][t]{9.6mm}{\includegraphics[width=9.6mm,bb=0 0 128 128]{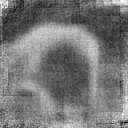} \newline \scriptsize{hedge\-hog}}
            \parbox[c][16.0mm][t]{9.6mm}{\includegraphics[width=9.6mm,bb=0 0 128 128]{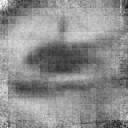} \newline \scriptsize{hel\-i-cop\-ter}}
            \parbox[c][16.0mm][t]{9.6mm}{\includegraphics[width=9.6mm,bb=0 0 128 128]{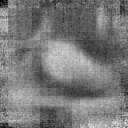} \newline \scriptsize{ibis}}
            \parbox[c][16.0mm][t]{9.6mm}{\includegraphics[width=9.6mm,bb=0 0 128 128]{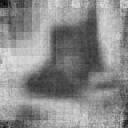} \newline \scriptsize{inline skate}}
            \parbox[c][16.0mm][t]{9.6mm}{\includegraphics[width=9.6mm,bb=0 0 128 128]{./data/classifier/joshua_tree_08_5.jpg} \newline \scriptsize{joshua tree}}
            \parbox[c][16.0mm][t]{9.6mm}{\includegraphics[width=9.6mm,bb=0 0 128 128]{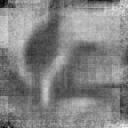} \newline \scriptsize{kan\-ga\-roo}}
            \parbox[c][16.0mm][t]{9.6mm}{\includegraphics[width=9.6mm,bb=0 0 128 128]{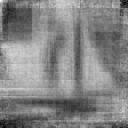} \newline \scriptsize{ketch}}
            \parbox[c][16.0mm][t]{9.6mm}{\includegraphics[width=9.6mm,bb=0 0 128 128]{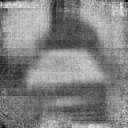} \newline \scriptsize{lamp}}
            \parbox[c][16.0mm][t]{9.6mm}{\includegraphics[width=9.6mm,bb=0 0 128 128]{./data/classifier/laptop_08_4.jpg} \newline \scriptsize{laptop}}
            \parbox[c][16.0mm][t]{9.6mm}{\includegraphics[width=9.6mm,bb=0 0 128 128]{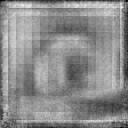} \newline \scriptsize{leopards}}
            \parbox[c][16.0mm][t]{9.6mm}{\includegraphics[width=9.6mm,bb=0 0 128 128]{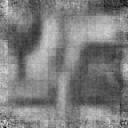} \newline \scriptsize{llama}}
            \parbox[c][16.0mm][t]{9.6mm}{\includegraphics[width=9.6mm,bb=0 0 128 128]{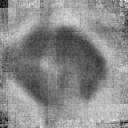} \newline \scriptsize{lobster}}
            \parbox[c][16.0mm][t]{9.6mm}{\includegraphics[width=9.6mm,bb=0 0 128 128]{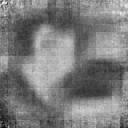} \newline \scriptsize{lotus}}
            \parbox[c][16.0mm][t]{9.6mm}{\includegraphics[width=9.6mm,bb=0 0 128 128]{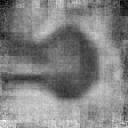} \newline \scriptsize{man\-do\-lin}}
            \parbox[c][16.0mm][t]{9.6mm}{\includegraphics[width=9.6mm,bb=0 0 128 128]{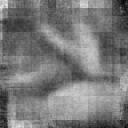} \newline \scriptsize{mayfly}}
            \parbox[c][16.0mm][t]{9.6mm}{\includegraphics[width=9.6mm,bb=0 0 128 128]{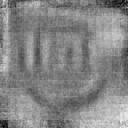} \newline \scriptsize{menorah}}
            \parbox[c][16.0mm][t]{9.6mm}{\includegraphics[width=9.6mm,bb=0 0 128 128]{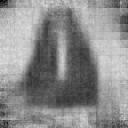} \newline \scriptsize{met\-ro\-nome}}
            \parbox[c][16.0mm][t]{9.6mm}{\includegraphics[width=9.6mm,bb=0 0 128 128]{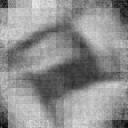} \newline \scriptsize{minaret}}
            \parbox[c][16.0mm][t]{9.6mm}{\includegraphics[width=9.6mm,bb=0 0 128 128]{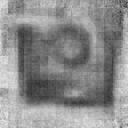} \newline \scriptsize{motor\-bikes}}
            \parbox[c][16.0mm][t]{9.6mm}{\includegraphics[width=9.6mm,bb=0 0 128 128]{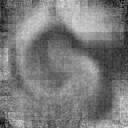} \newline \scriptsize{nautilus}}
            \parbox[c][16.0mm][t]{9.6mm}{\includegraphics[width=9.6mm,bb=0 0 128 128]{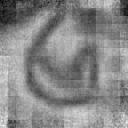} \newline \scriptsize{octopus}}
            \parbox[c][16.0mm][t]{9.6mm}{\includegraphics[width=9.6mm,bb=0 0 128 128]{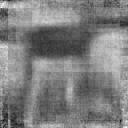} \newline \scriptsize{okapi}}
            \parbox[c][16.0mm][t]{9.6mm}{\includegraphics[width=9.6mm,bb=0 0 128 128]{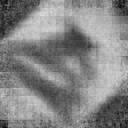} \newline \scriptsize{pagoda}}
            \parbox[c][16.0mm][t]{9.6mm}{\includegraphics[width=9.6mm,bb=0 0 128 128]{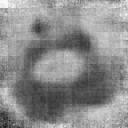} \newline \scriptsize{panda}}
            \parbox[c][16.0mm][t]{9.6mm}{\includegraphics[width=9.6mm,bb=0 0 128 128]{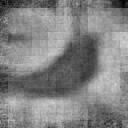} \newline \scriptsize{pigeon}}
            \parbox[c][16.0mm][t]{9.6mm}{\includegraphics[width=9.6mm,bb=0 0 128 128]{./data/classifier/pizza_08_3.jpg} \newline \scriptsize{pizza}}
            \parbox[c][16.0mm][t]{9.6mm}{\includegraphics[width=9.6mm,bb=0 0 128 128]{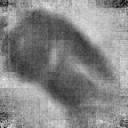} \newline \scriptsize{platypus}}
            \parbox[c][16.0mm][t]{9.6mm}{\includegraphics[width=9.6mm,bb=0 0 128 128]{./data/classifier/pyramid_08_2.jpg} \newline \scriptsize{pyramid}}
            \parbox[c][16.0mm][t]{9.6mm}{\includegraphics[width=9.6mm,bb=0 0 128 128]{./data/classifier/revolver_08_1.jpg} \newline \scriptsize{revolver}}
            \parbox[c][16.0mm][t]{9.6mm}{\includegraphics[width=9.6mm,bb=0 0 128 128]{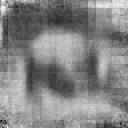} \newline \scriptsize{rhino}}
            \parbox[c][16.0mm][t]{9.6mm}{\includegraphics[width=9.6mm,bb=0 0 128 128]{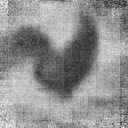} \newline \scriptsize{rooster}}
            \parbox[c][16.0mm][t]{9.6mm}{\includegraphics[width=9.6mm,bb=0 0 128 128]{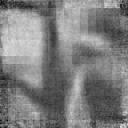} \newline \scriptsize{saxo\-phone}}
            \parbox[c][16.0mm][t]{9.6mm}{\includegraphics[width=9.6mm,bb=0 0 128 128]{./data/classifier/schooner_08_3.jpg} \newline \scriptsize{schooner}}
            \parbox[c][16.0mm][t]{9.6mm}{\includegraphics[width=9.6mm,bb=0 0 128 128]{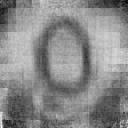} \newline \scriptsize{scissors}}
            \parbox[c][16.0mm][t]{9.6mm}{\includegraphics[width=9.6mm,bb=0 0 128 128]{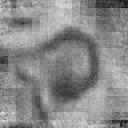} \newline \scriptsize{scorpion}}
            \parbox[c][16.0mm][t]{9.6mm}{\includegraphics[width=9.6mm,bb=0 0 128 128]{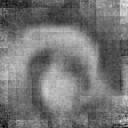} \newline \scriptsize{sea horse}}
            \parbox[c][16.0mm][t]{9.6mm}{\includegraphics[width=9.6mm,bb=0 0 128 128]{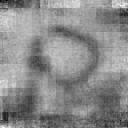} \newline \scriptsize{snoopy}}
            \parbox[c][16.0mm][t]{9.6mm}{\includegraphics[width=9.6mm,bb=0 0 128 128]{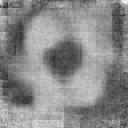} \newline \scriptsize{soccer ball}}
            \parbox[c][16.0mm][t]{9.6mm}{\includegraphics[width=9.6mm,bb=0 0 128 128]{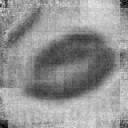} \newline \scriptsize{stapler}}
            \parbox[c][16.0mm][t]{9.6mm}{\includegraphics[width=9.6mm,bb=0 0 128 128]{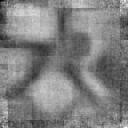} \newline \scriptsize{starfish}}
            \parbox[c][16.0mm][t]{9.6mm}{\includegraphics[width=9.6mm,bb=0 0 128 128]{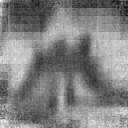} \newline \scriptsize{stego\-saurus}}
            \parbox[c][16.0mm][t]{9.6mm}{\includegraphics[width=9.6mm,bb=0 0 128 128]{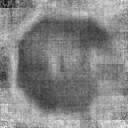} \newline \scriptsize{stop sign}}
            \parbox[c][16.0mm][t]{9.6mm}{\includegraphics[width=9.6mm,bb=0 0 128 128]{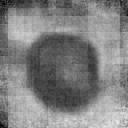} \newline \scriptsize{straw\-berry}}
            \parbox[c][16.0mm][t]{9.6mm}{\includegraphics[width=9.6mm,bb=0 0 128 128]{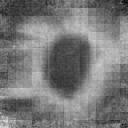} \newline \scriptsize{sun\-flower}}
            \parbox[c][16.0mm][t]{9.6mm}{\includegraphics[width=9.6mm,bb=0 0 128 128]{./data/classifier/tick_08_1.jpg} \newline \scriptsize{tick}}
            \parbox[c][16.0mm][t]{9.6mm}{\includegraphics[width=9.6mm,bb=0 0 128 128]{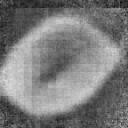} \newline \scriptsize{trilobite}}
            \parbox[c][16.0mm][t]{9.6mm}{\includegraphics[width=9.6mm,bb=0 0 128 128]{./data/classifier/umbrella_08_2.jpg} \newline \scriptsize{um\-brel\-la}}
            \parbox[c][16.0mm][t]{9.6mm}{\includegraphics[width=9.6mm,bb=0 0 128 128]{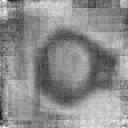} \newline \scriptsize{watch}}
            \parbox[c][16.0mm][t]{9.6mm}{\includegraphics[width=9.6mm,bb=0 0 128 128]{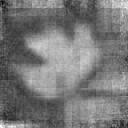} \newline \scriptsize{water lilly}}
            \parbox[c][16.0mm][t]{9.6mm}{\includegraphics[width=9.6mm,bb=0 0 128 128]{./data/classifier/wheelchair_08_5.jpg} \newline \scriptsize{wheel\-chair}}
            \parbox[c][16.0mm][t]{9.6mm}{\includegraphics[width=9.6mm,bb=0 0 128 128]{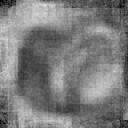} \newline \scriptsize{wild cat}}
            \parbox[c][16.0mm][t]{9.6mm}{\includegraphics[width=9.6mm,bb=0 0 128 128]{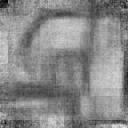} \newline \scriptsize{windsor chair}}
            \parbox[c][16.0mm][t]{9.6mm}{\includegraphics[width=9.6mm,bb=0 0 128 128]{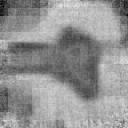} \newline \scriptsize{wrench}}
            \parbox[c][16.0mm][t]{9.6mm}{\includegraphics[width=9.6mm,bb=0 0 128 128]{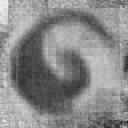} \newline \scriptsize{yin yang}}
            \parbox[c][16.0mm][t]{9.6mm}{\includegraphics[width=9.6mm,bb=0 0 128 128]{./data/blank.jpg} \newline \scriptsize{}}
            \caption{Visualized object classifiers.}
            \label{fig:appendix_classifier}
        \end{figure*}

                Examples of visualized classifiers are shown in Fig.~\ref{fig:classifier}. All results are shown in Fig.~\ref{fig:appendix_classifier}. We can understand that the pizza classifier strongly captures a circular shape. It is interesting that a clear horizontal line exists in the image of the Joshua tree classifier. This fact suggests that a horizon is a strong clue for the classifier. It is also surprising that our very simple method produced well-identifiable images, which indicates that the inner class variation of images in Caltech $101$ Dataset is extremely limited.

        \subsection{Image Generation from Natural Sentences}
            \label{sec:application_sentence}

            Automatic caption generation is studied extensively in recent years~\cite{ushiku2012efficient,vinyals2014show}. Caption-generation systems capture the relation between images and their captions using image caption dataset. Here, if such a relation is captured, then it is possible to generate not only natural sentences from images, but also images from natural sentences. 

            Recent caption-generation systems use Convolutional Neural Networks for image feature extraction. However, they are not invertible. Our work enables inversion of BoVW. In this section, we use BoVW as an intermediate feature for image generation from sentences.
            
            \subsubsection{Method}

                To generate images from natural sentences, we decompose a sentence into words, convert each word to a BoVW, average them to get one BoVW, convert this float-value BoVW into integer-value BoVW in the same manner described in Section~\ref{sec:application_classifier}, and generate an image from the obtained BoVW.

                The method to convert a word $w$ into a BoVW is the following. We use an image caption dataset $I$. If $w$ appears $n$ times in the caption of $i$-th image in $I$, then we set $s_i = n$. If the visual word $v_j$ appears $m$ times in the BoVW of $i$-th image, then we set $t_{ij}=m$. Then we compute correlation coefficient $u_j$ between $s_i$ and $t_{ij}$ for all $j$. Finally, we concatenate $u_j$ and treat it as BoVW for $w$.

            \subsubsection{Experiment}

                We used the Pascal Sentence Dataset~\cite{rashtchian2010collecting} as an image caption dataset. This dataset includes one thousand images with five captions each. We concatenate five captions and treat them as one caption. Additionally, we only used nouns for conversion. We used RGBSIFT as local descriptors.

\begin{figure*}
                    \centering
                    \subfloat[Boat on a beach.]{\makebox[35mm][c]{\includegraphics[width=13.4mm,bb=0 0 128 128]{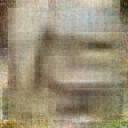}}}
                    \subfloat[Bottle on a cliff.]{\makebox[35mm][c]{\includegraphics[width=13.4mm,bb=0 0 128 128]{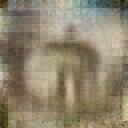}}}
                    \subfloat[Bus on a field.]{\makebox[35mm][c]{\includegraphics[width=13.4mm,bb=0 0 128 128]{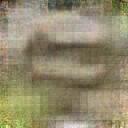}}}
                    \subfloat[Monitor on a coast.]{\makebox[35mm][c]{\includegraphics[width=13.4mm,bb=0 0 128 128]{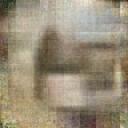}}}
                    \subfloat[Plane over a mountain.]{\makebox[35mm][c]{\includegraphics[width=13.4mm,bb=0 0 128 128]{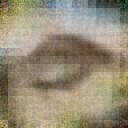}}}
                   \caption{Images generated from natural sentences.}
                   \label{fig:from_sentences}
                \end{figure*}

                Examples of generated images are shown in Fig.~\ref{fig:from_sentences}. They are not so clear, but there are slight correspondences to the input sentences. For example, the image of Fig.~\ref{fig:from_sentences}(b) has a vertically long object corresponding to "bottle", Fig.~\ref{fig:from_sentences}(c) has a horizontally long object corresponding to "bus". Images of "field" have many green parts but "coast" and "beach" do not.

                In this setting, several sentences are translated to completely nonsense images. Our method is probably too simple for this complicated task. The image caption dataset is small against the variety of images and sentences. However, the fact that this simple method and small dataset can produce results such as those in Fig.~\ref{fig:from_sentences} indicates the future possibility of realizing image generation systems.

    \section{Discussion}
        \label{chap:discussion}

        Our method assumes dense sampling, single-scale sampling, and hard assignment of local descriptors. These limitations prevent application of our method to more advanced BoVW based features. To relax these assumptions, additional terms must be added to the objective function to estimate extra parameters such as orientation, scale, and soft assignment of descriptors.

        Our method also presumes that the size and the extraction step of local descriptors are available. These assumptions are appropriate when the users intend to analyze BoVW to improve their own systems, or when developing novel applications. However, because of this limitation, it is impossible to reconstruct images from BoVWs which are extracted by unknown systems. 

        As described in Section~\ref{sec:experiment_bovw}, although human evaluation clearly indicates strong superiority of our approach to others, the difference is small in view of XCORR$n$. Because human evaluation costs are high, and because they take much time, and are subject to the selection of participants, development of more reliable evaluation metrics will be necessary to accelerate future research in this field.

        In this work, we did not use Spatial Pyramid~\cite{lazebnik2006beyond}. However, if the Spatial Pyramid is used, possible positions of visual words are greatly limited, which might produce more clear reconstructions within much less optimization time.

    \section{Conclusion}
        \label{chap:conclusion}

        As described herein, we addressed a problem of reconstructing original images from Bag-of-Visual-Word feature. We decomposed this novel task into estimation of optimal arrangement of visual words and generation of images from them. We specifically emphasized the former subproblem and proposed a method which acts like a jigsaw puzzle solver using statistics of visual words in an image dataset. 

        Results of the experiments revealed the following. 1) Our method can reconstruct images from BoVW. 2) In the estimation of the original arrangement of visual words, it is effective to consider both the naturalness of preference of global position and naturalness of their local adjacencies. 3) Quantization error, which emerges at the coding step, is not problematic to generate image patches, but it strongly affects the estimation of geometric information of visual words. 4) Our optimization method offers a correct reconstruction, although it is time consuming. 5) Our method is applicable to local descriptors of arbitrary kinds.

        Additionally, we applied our method to three other tasks: 1) image morphing via BoVW, 2) visualization of object classifiers, and 3) image generation from natural sentences. Although the proposed methods are simple, they produced promising results.
        
        We believe that image reconstruction from features is an extremely important theme because it is useful for both feature analysis and interesting but challenging applications. We expect our work related to well-used Bag-of-Visual-Words invokes more intensive research in this field.


        \renewcommand{\baselinestretch}{0.9}
        \renewcommand{\baselinestretch}{1.0}




    \bibliographystyle{IEEEtran}
    \bibliography{tpami}

\begin{thebibliography}{10}
\providecommand{\url}[1]{#1}
\csname url@samestyle\endcsname
\providecommand{\newblock}{\relax}
\providecommand{\bibinfo}[2]{#2}
\providecommand{\BIBentrySTDinterwordspacing}{\spaceskip=0pt\relax}
\providecommand{\BIBentryALTinterwordstretchfactor}{4}
\providecommand{\BIBentryALTinterwordspacing}{\spaceskip=\fontdimen2\font plus
\BIBentryALTinterwordstretchfactor\fontdimen3\font minus
  \fontdimen4\font\relax}
\providecommand{\BIBforeignlanguage}[2]{{%
\expandafter\ifx\csname l@#1\endcsname\relax
\typeout{** WARNING: IEEEtran.bst: No hyphenation pattern has been}%
\typeout{** loaded for the language `#1'. Using the pattern for}%
\typeout{** the default language instead.}%
\else
\language=\csname l@#1\endcsname
\fi
#2}}
\providecommand{\BIBdecl}{\relax}
\BIBdecl

\bibitem{weinzaepfel2011reconstructing}
P.~Weinzaepfel, H.~J{\'e}gou, and P.~P{\'e}rez, ``Reconstructing an image from
  its local descriptors,'' in \emph{Proc. IEEE Conf. Computer Vision and
  Pattern Recognition}, 2011.

\bibitem{d2012beyond}
E.~d'Angelo, A.~Alahi, and P.~Vandergheynst, ``Beyond bits: Reconstructing
  images from local binary descriptors,'' in \emph{Proc. IEEE Conf. Pattern
  Recognition}, 2012.

\bibitem{vondrick2012inverting}
C.~Vondrick, A.~Khosla, T.~Malisiewicz, and A.~Torralba, ``Hoggles: Visualizing
  object detection features,'' in \emph{Proc. IEEE Int'l Conf. Computer
  Vision}, 2013.

\bibitem{zeiler2013visualizing}
M.~D. Zeiler and R.~Fergus, ``Visualizing and understanding convolutional
  neural networks,'' in \emph{Proc. European Conf. Computer Vision}, 2014.

\bibitem{ILSVRCarxiv14}
O.~Russakovsky, J.~Deng, H.~Su, J.~Krause, S.~Satheesh, S.~Ma, Z.~Huang,
  A.~Karpathy, A.~Khosla, M.~Bernstein, A.~C. Berg, and L.~Fei-Fei, ``Imagenet
  large scale visual recognition challenge,'' \emph{Int'l J. Computer Vision},
  pp. 1--42, 2015.

\bibitem{chen2014inferring}
C.-Y. Chen and K.~Grauman, ``Inferring unseen views of people,'' in \emph{Proc.
  IEEE Conf. Computer Vision and Pattern Recognition}, 2014.

\bibitem{zen2009statistical}
H.~Zen, K.~Tokuda, and A.~W. Black, ``Statistical parametric speech
  synthesis,'' \emph{Speech Communication}, vol.~51, no.~11, pp. 1039--1064,
  2009.

\bibitem{stylianou1998continuous}
Y.~Stylianou, O.~Capp{\'e}, and E.~Moulines, ``Continuous probabilistic
  transform for voice conversion,'' \emph{IEEE Trans. Speech and Audio
  Processing}, vol.~6, no.~2, pp. 131--142, 1998.

\bibitem{sanchez2013image}
J.~S{\'a}nchez, F.~Perronnin, T.~Mensink, and J.~Verbeek, ``Image
  classification with the fisher vector: Theory and practice,'' \emph{Int'l J.
  Computer Vision}, vol. 105, no.~3, pp. 222--245, 2013.

\bibitem{zhou2010image}
X.~Zhou, K.~Yu, T.~Zhang, and T.~S. Huang, ``Image classifcation using
  super-vector coding of local image descriptors,'' in \emph{Proc. European
  Conf. Computer Vision}, 2010.

\bibitem{Kato_CVPR_2014}
H.~Kato and T.~Harada, ``Image reconstruction from bag-of-visual-words,'' in
  \emph{Proc. IEEE Conf. Computer Vision and Pattern Recognition}, 2014.

\bibitem{lowe2004distinctive}
D.~G. Lowe, ``Distinctive image features from scale-invariant keypoints,''
  \emph{Int'l J. Computer Vision}, vol.~60, no.~2, pp. 91--110, 2004.

\bibitem{perez2003poisson}
P.~P{\'e}rez, M.~Gangnet, and A.~Blake, ``Poisson image editing,'' \emph{ACM
  Trans. Graphics}, vol.~22, no.~3, pp. 313--318, 2003.

\bibitem{calonder2010brief}
M.~Calonder, V.~Lepetit, C.~Strecha, and P.~Fua, ``Brief: Binary robust
  independent elementary features,'' in \emph{Proc. European Conf. Computer
  Vision}, 2010.

\bibitem{alahi2012freak}
A.~Alahi, R.~Ortiz, and P.~Vandergheynst, ``Freak: Fast retina keypoint,'' in
  \emph{Proc. IEEE Conf. Computer Vision and Pattern Recognition}, 2012.

\bibitem{dalal2005histograms}
N.~Dalal and B.~Triggs, ``Histograms of oriented gradients for human
  detection,'' in \emph{Proc. IEEE Conf. Computer Vision and Pattern
  Recognition}, 2005.

\bibitem{lecun1998gradient}
Y.~LeCun, L.~Bottou, Y.~Bengio, and P.~Haffner, ``Gradient-based learning
  applied to document recognition,'' vol.~86, no.~11, pp. 2278--2324, 1998.

\bibitem{zeiler2010deconvolutional}
M.~D. Zeiler, D.~Krishnan, G.~W. Taylor, and R.~Fergus, ``Deconvolutional
  networks,'' in \emph{Proc. IEEE Conf. Computer Vision and Pattern
  Recognition}, 2010.

\bibitem{mahendran2014understanding}
A.~Mahendran and A.~Vedaldi, ``Understanding deep image representations by
  inverting them,'' in \emph{Proc. IEEE Conf. Computer Vision and Pattern
  Recognition}, 2015.

\bibitem{johnson2006semantic}
M.~Johnson, G.~J. Brostow, J.~Shotton, O.~Arandjelovic, V.~Kwatra, and
  R.~Cipolla, ``Semantic photo synthesis,'' \emph{Computer Graphics Forum},
  vol.~25, no.~3, pp. 407--413, 2006.

\bibitem{chen2009sketch2photo}
T.~Chen, M.-M. Cheng, P.~Tan, A.~Shamir, and S.-M. Hu, ``Sketch2photo: internet
  image montage,'' \emph{ACM Trans. Graphics}, vol.~28, no.~5, p. 124, 2009.

\bibitem{erhan2009visualizing}
D.~Erhan, Y.~Bengio, A.~Courville, and P.~Vincent, ``Visualizing higher-layer
  features of a deep network,'' in \emph{Proc. ICML Workshop Learning Feature
  Hierarchies}, 2009.

\bibitem{bengio2012better}
Y.~Bengio, G.~Mesnil, Y.~Dauphin, and S.~Rifai, ``Better mixing via deep
  representations,'' in \emph{Proc. Int'l Conf. Machine Learning}, 2013.

\bibitem{kingma2013auto}
D.~P. Kingma and M.~Welling, ``Auto-encoding variational bayes,'' in
  \emph{Proc. Int'l Conf. Learning Representations}, 2014.

\bibitem{graves2013generating}
A.~Graves, ``Generating sequences with recurrent neural networks,''
  \emph{arXiv:1308.0850}, 2013.

\bibitem{gregor2015draw}
K.~Gregor, I.~Danihelka, A.~Graves, and D.~Wierstra, ``Draw: A recurrent neural
  network for image generation,'' in \emph{Proc. Int'l Conf. Machine Learning},
  2015.

\bibitem{krizhevsky2009learning}
A.~Krizhevsky and G.~Hinton, ``Learning multiple layers of features from tiny
  images,'' \emph{Univ. Toronto, Tech. Rep}, vol.~1, no.~4, p.~7, 2009.

\bibitem{sivic2003video}
J.~Sivic and A.~Zisserman, ``Video google: A text retrieval approach to object
  matching in videos,'' in \emph{Proc. IEEE Int'l Conf. Computer Vision}, 2003.

\bibitem{csurka2004visual}
G.~Csurka, C.~Dance, L.~Fan, J.~Willamowski, and C.~Bray, ``Visual
  categorization with bags of keypoints,'' in \emph{Proc. ECCV Workshop
  Statistical Learning in Computer Vision}, 2004.

\bibitem{harris1988combined}
C.~Harris and M.~Stephens, ``A combined corner and edge detector.'' in
  \emph{Proc. Alvey Vision Conference}, 1988.

\bibitem{nowak2006sampling}
E.~Nowak, F.~Jurie, and B.~Triggs, ``Sampling strategies for bag-of-features
  image classification,'' in \emph{Proc. European Conf. Computer Vision}, 2006.

\bibitem{gordoa2012leveraging}
A.~Gordoa, J.~A. Rodr{\'\i}guez-Serrano, F.~Perronnin, and E.~Valveny,
  ``Leveraging category-level labels for instance-level image retrieval,'' in
  \emph{Proc. IEEE Conf. Computer Vision and Pattern Recognition}, 2012.

\bibitem{winn2005object}
J.~Winn, A.~Criminisi, and T.~Minka, ``Object categorization by learned
  universal visual dictionary,'' in \emph{Proc. IEEE Int'l Conf. Computer
  Vision}, 2005.

\bibitem{perronnin2006adapted}
F.~Perronnin, C.~Dance, G.~Csurka, and M.~Bressan, ``Adapted vocabularies for
  generic visual categorization,'' in \emph{Proc. European Conf. Computer
  Vision}, 2006.

\bibitem{van2008kernel}
J.~C. van Gemert, J.-M. Geusebroek, C.~J. Veenman, and A.~W. Smeulders,
  ``Kernel codebooks for scene categorization,'' in \emph{Proc. European Conf.
  Computer Vision}, 2008.

\bibitem{yang2009linear}
J.~Yang, K.~Yu, Y.~Gong, and T.~Huang, ``Linear spatial pyramid matching using
  sparse coding for image classification,'' in \emph{Proc. IEEE Conf. Computer
  Vision and Pattern Recognition}, 2009.

\bibitem{wang2010locality}
J.~Wang, J.~Yang, K.~Yu, F.~Lv, T.~Huang, and Y.~Gong, ``Locality-constrained
  linear coding for image classification,'' in \emph{Proc. IEEE Conf. Computer
  Vision and Pattern Recognition}, 2010.

\bibitem{boureau2010learning}
Y.-L. Boureau, F.~Bach, Y.~LeCun, and J.~Ponce, ``Learning mid-level features
  for recognition,'' in \emph{Proc. IEEE Conf. Computer Vision and Pattern
  Recognition}, 2010.

\bibitem{boureau2010theoretical}
Y.-L. Boureau, J.~Ponce, and Y.~LeCun, ``A theoretical analysis of feature
  pooling in visual recognition,'' in \emph{Proc. Int'l Conf. Machine
  Learning}, 2010.

\bibitem{lazebnik2006beyond}
S.~Lazebnik, C.~Schmid, and J.~Ponce, ``Beyond bags of features: Spatial
  pyramid matching for recognizing natural scene categories,'' in \emph{Proc.
  IEEE Conf. Computer Vision and Pattern Recognition}, 2006.

\bibitem{jegou2010aggregating}
H.~J{\'e}gou, M.~Douze, C.~Schmid, and P.~P{\'e}rez, ``Aggregating local
  descriptors into a compact image representation,'' in \emph{Proc. IEEE Conf.
  Computer Vision and Pattern Recognition}, 2010.

\bibitem{chatfield2011devil}
K.~Chatfield, V.~Lempitsky, A.~Vedaldi, and A.~Zisserman, ``The devil is in the
  details: an evaluation of recent feature encoding methods,'' in \emph{Proc.
  British Machine Vision Conf.}, 2011.

\bibitem{cho2010probabilistic}
T.~S. Cho, S.~Avidan, and W.~T. Freeman, ``A probabilistic image jigsaw puzzle
  solver,'' in \emph{Proc. IEEE Conf. Computer Vision and Pattern Recognition},
  2010.

\bibitem{pomeranz2011fully}
D.~Pomeranz, M.~Shemesh, and O.~Ben-Shahar, ``A fully automated greedy square
  jigsaw puzzle solver,'' in \emph{Proc. IEEE Conf. Computer Vision and Pattern
  Recognition}, 2011.

\bibitem{gallagher2012jigsaw}
A.~C. Gallagher, ``Jigsaw puzzles with pieces of unknown orientation,'' in
  \emph{Proc. IEEE Conf. Computer Vision and Pattern Recognition}, 2012.

\bibitem{sholomon2013genetic}
D.~Sholomon, O.~David, and N.~S. Netanyahu, ``A genetic algorithm-based solver
  for very large jigsaw puzzles,'' in \emph{Proc. IEEE Conf. Computer Vision
  and Pattern Recognition}, 2013.

\bibitem{pearl1988probabilistic}
J.~Pearl, \emph{Probabilistic Reasoning in Intelligent Systems: Networks of
  Plausible Inference}, 1988.

\bibitem{boykov2001fast}
Y.~Boykov, O.~Veksler, and R.~Zabih, ``Fast approximate energy minimization via
  graph cuts,'' \emph{IEEE Trans. Pattern Analysis and Machine Intelligence},
  vol.~23, no.~11, pp. 1222--1239, 2001.

\bibitem{kolmogorov2006convergent}
V.~Kolmogorov, ``Convergent tree-reweighted message passing for energy
  minimization,'' \emph{IEEE Trans. Pattern Analysis and Machine Intelligence},
  vol.~28, no.~10, pp. 1568--1583, 2006.

\bibitem{cho2008patch}
T.~S. Cho, M.~Butman, S.~Avidan, and W.~T. Freeman, ``The patch transform and
  its applications to image editing,'' in \emph{Proc. IEEE Conf. Computer
  Vision and Pattern Recognition}, 2008.

\bibitem{koopmans1957assignment}
T.~C. Koopmans and M.~Beckmann, ``Assignment problems and the location of
  economic activities,'' \emph{Econometrica}, vol.~25, no.~1, pp. 53--76, 1957.

\bibitem{lawler1963quadratic}
E.~L. Lawler, ``The quadratic assignment problem,'' \emph{Management Science},
  vol.~9, no.~4, pp. 586--599, 1963.

\bibitem{loiola2007survey}
E.~M. Loiola, N.~M.~M. De~Abreu, P.~O. Boaventura-Netto, P.~Hahn, and
  T.~Querido, ``A survey for the quadratic assignment problem,'' \emph{European
  J. Operational Research}, vol. 176, no.~2, pp. 657--690, 2007.

\bibitem{drezner2008extensive}
Z.~Drezner, ``Extensive experiments with hybrid genetic algorithms for the
  solution of the quadratic assignment problem,'' \emph{Computers \& Operations
  Research}, vol.~35, no.~3, pp. 717--736, 2008.

\bibitem{drezner2003new}
------, ``A new genetic algorithm for the quadratic assignment problem,''
  \emph{INFORMS J. Computing}, vol.~15, no.~3, pp. 320--330, 2003.

\bibitem{fei2007learning}
L.~Fei-Fei, R.~Fergus, and P.~Perona, ``Learning generative visual models from
  few training examples: An incremental bayesian approach tested on 101 object
  categories,'' \emph{Computer Vision and Image Understanding}, vol. 106,
  no.~1, pp. 59--70, 2007.

\bibitem{harada2012graphical}
T.~Harada and Y.~Kuniyoshi, ``Graphical gaussian vector for image
  categorization,'' in \emph{Proc. Neural Information Processing Systems},
  2012.

\bibitem{van2008comparison}
K.~E. van~de Sande, T.~Gevers, and C.~G. Snoek, ``A comparison of color
  features for visual concept classification,'' in \emph{Proc. Int'l Conf.
  Content-based Image and Video Retrieval}, 2008.

\bibitem{ojala1996comparative}
T.~Ojala, M.~Pietik{\"a}inen, and D.~Harwood, ``A comparative study of texture
  measures with classification based on featured distributions,'' \emph{Pattern
  Recognition}, vol.~29, no.~1, pp. 51--59, 1996.

\bibitem{nguyen2014deep}
A.~Nguyen, J.~Yosinski, and J.~Clune, ``Deep neural networks are easily fooled:
  High confidence predictions for unrecognizable images,'' in \emph{Proc. IEEE
  Conf. Computer Vision and Pattern Recognition}, 2015.

\bibitem{ushiku2012efficient}
Y.~Ushiku, T.~Harada, and Y.~Kuniyoshi, ``Efficient image annotation for
  automatic sentence generation,'' in \emph{ACM Conf. Multimedia}, 2012.

\bibitem{vinyals2014show}
O.~Vinyals, A.~Toshev, S.~Bengio, and D.~Erhan, ``Show and tell: A neural image
  caption generator,'' in \emph{Proc. IEEE Conf. Computer Vision and Pattern
  Recognition}, 2015.

\bibitem{rashtchian2010collecting}
C.~Rashtchian, P.~Young, M.~Hodosh, and J.~Hockenmaier, ``Collecting image
  annotations using amazon's mechanical turk,'' in \emph{Proc. NAACL HLT
  Workshop Creating Speech and Language Data with Amazon's Mechanical Turk},
  2010.

\end{thebibliography}


    \begin{IEEEbiography}[{\includegraphics[width=1in,height=1.25in,clip,keepaspectratio,bb=0 0 1500 1890]{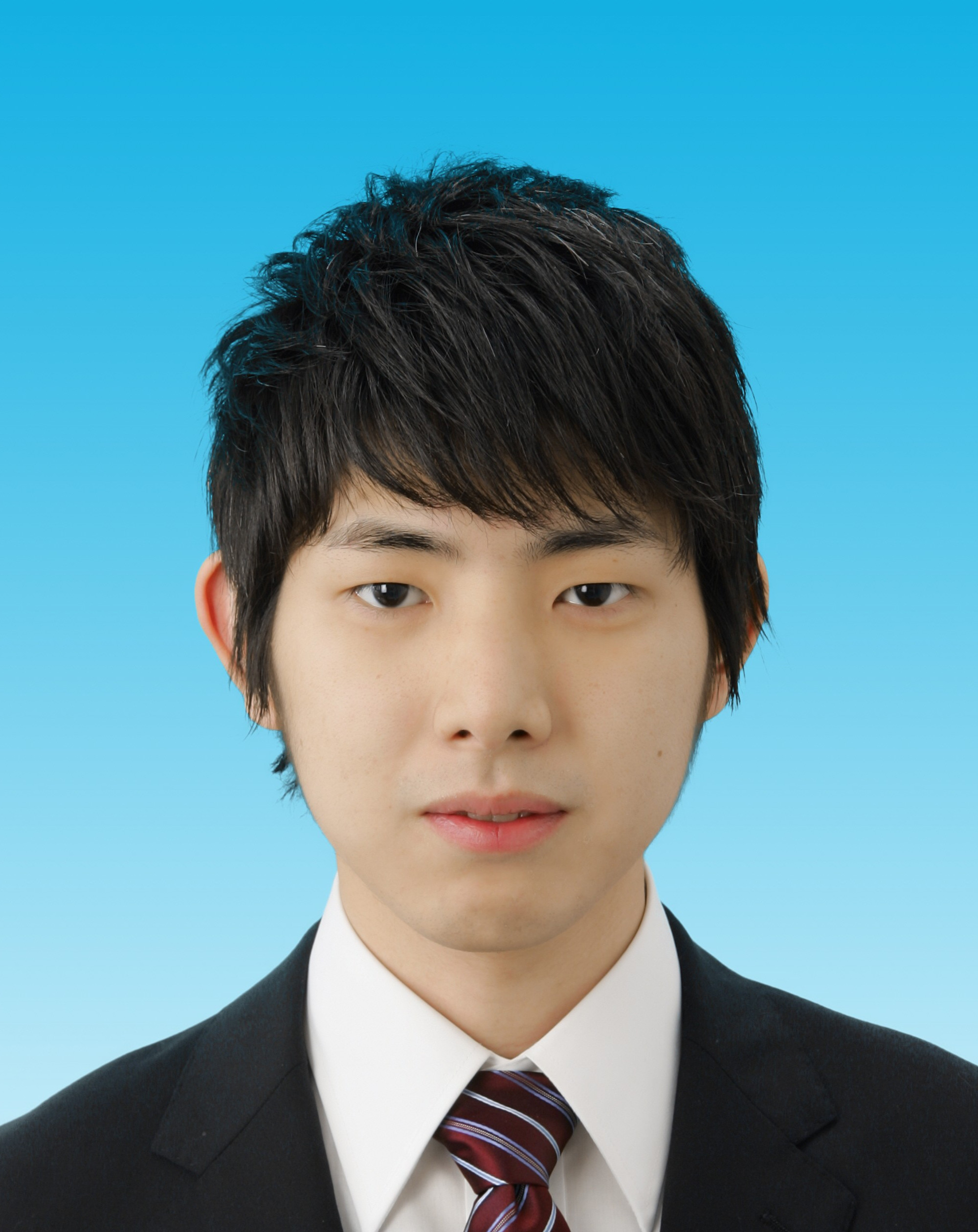}}]{Hiroharu Kato}
        received the BS degree (2012) in engineering and the MS degree (2014) in information science and technology from The University of Tokyo, Japan. He is currently a research engineer at Sony Corporation. His research interests include multimedia processing, artificial intelligence, and their industrial applications.
    \end{IEEEbiography}

    \begin{IEEEbiography}[{\includegraphics[width=1in,height=1.25in,clip,keepaspectratio,bb=0 0 1638 2047]{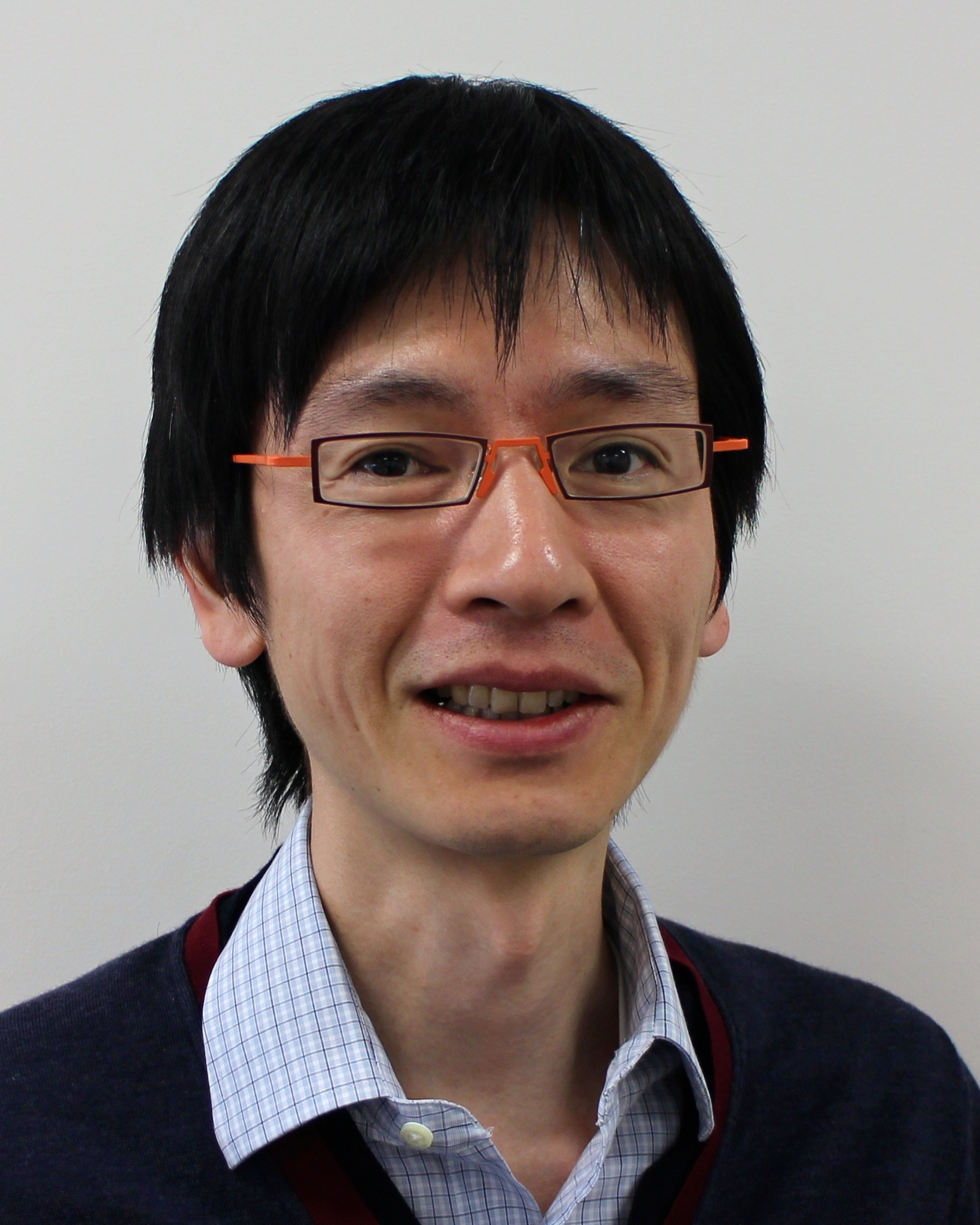}}]{Tatsuya Harada}
        received PhD degree (2001) in mechanical engineering from The University of Tokyo, Japan. He is currently a Professor at the Department of Mechano-Informatics, School of Information Science and Technology, The University of Tokyo, Japan. His research interests include real world intelligent system, large scale visual recognition, as well as intelligent robot.
    \end{IEEEbiography}

\end{document}